\documentclass[11pt]{article}

\usepackage{acl}

\usepackage[utf8]{inputenc}
\usepackage[T1]{fontenc}
\usepackage{times}       
\usepackage{latexsym}
\usepackage{inconsolata} 
\usepackage{amsmath,amssymb,amsfonts}
\usepackage{graphicx}
\usepackage{booktabs}
\usepackage{multirow}
\usepackage{algorithm}
\usepackage{algorithmic}
\usepackage{float}
\usepackage{xcolor}
\usepackage{tikz}
\usepackage{pgfplots}
\pgfplotsset{compat=1.18}
\usetikzlibrary{positioning, calc}
\usepackage{subcaption}
\usepackage{enumitem}
\usepackage{placeins}  
\newcommand{\sys}{\textsc{AURA}}
\newcommand{\envagent}{Environment Agent}

\title{\sys{}: Intent-Directed Probing for Implicit-Need Surfacing \\ in Situated LLM Agents}

\author{%
  Yang Li, \; Jiaxiang Liu, \; Jiang Cai, \; Mingkun Xu\thanks{\;Corresponding author.} \\
  Guangdong Institute of Intelligence Science and Technology, Zhuhai, China \\
  \texttt{\{liyang, liujiaxiang, caijiang, xumingkun\}@gdiist.cn} \\
}

\begin{document}
\renewcommand{\thefootnote}{\fnsymbol{footnote}}
\maketitle
\renewcommand{\thefootnote}{\arabic{footnote}}
\setcounter{footnote}{0}

\begin{abstract}
A situated query like ``where is Lin Wei?'' often encodes more than its literal content: the user may also want to know whether Lin Wei is free, in a good mood, or worth interrupting now. Standard tool-use agents answer the literal question and stop. \sys{} inserts an inference step between scene perception and tool use that produces an \texttt{IntentFrame}: a structured estimate of the implicit need with a scalar \emph{gap} score that controls per-query probe budget and tool selection. On a 100-query four-scene implicit-intent benchmark, \sys{} improves implicit-need coverage over ReAct-style probing ($\boldsymbol{\Delta{=}{+}0.07}$, $\boldsymbol{p{<}10^{-6}}$); three of four scenes are individually significant, the gain reproduces on a second backbone, and a prompt ablation attributes the lift to gap calibration rather than answer memorisation. A component ablation localises the gain further: the gap-derived budget ceiling alone outperforms the full architecture ($p{=}0.02$), and bin-wise calibration analysis ($r{=}{-}0.33$) identifies the gap score as a probe \emph{trigger} rather than a calibrated value estimator. On factual lookup the controller trades raw accuracy for $\mathbf{82\%}$ fewer probes and zero forbidden-tool violations on a privacy-sensitive slice; scope conditions are detailed in Limitations. Code, simulator, and benchmark are released at \url{https://github.com/innovation64/AURA}, with the implicit-intent benchmark also on the HuggingFace Hub (\texttt{innovation64/aura-implicit-intent}).

\end{abstract}

\section{Introduction}
\label{sec:intro}

LLM-based agents~\citep{xi2023rise, wang2024survey} have been deployed in social simulations~\citep{park2023generative} and software teams~\citep{hong2024metagpt}, but they often answer the user's \emph{literal} question about an environment while missing the \emph{implicit} information need behind it. Asking ``where is Lin Wei?'' may be a request for location, but it may also mean ``is she free to chat?''. Agents then decide with incomplete context, invent details instead of grounding in current state, and rarely translate environment state into what the user needs next.

Prior approaches answer this only partially. \textbf{ReAct}~\citep{yao2023react} interleaves reasoning and tool calls during answer generation, but tools fire only when the surface query explicitly demands them --- the loop has no step that asks ``what does the user actually want to know?''. \textbf{Plan-and-Solve}~\citep{wang2023plansolve} pre-plans tool calls from the literal query, with no mechanism to bridge to an implicit need that the literal query does not name. \textbf{Generative Agents}~\citep{park2023generative} inject all environment state as passive context but expose no control over which private state to surface for a given query. Each treats the surface query as if it fully specified what the user wants. \sys{} adds the missing step: implicit-need inference as an \emph{independent control variable}, a pre-tool routing decision that determines which private-state probes to issue and how many.

By \emph{situated} we mean structured environments where private state sits behind tool-mediated access: public context is observed passively, hidden state must be probed. This is narrower than the ``situated agent'' umbrella (embodied navigation, open-world exploration, sensor fusion), which we do not address.
We study this problem as implicit-intent inference over structured environment state. \sys{} introduces the \textbf{\envagent{}}: a pipeline (Perceive $\to$ Scene $\to$ Memory $\to$ Reason) plus an LLM-generated \texttt{IntentFrame} that estimates the gap between a user's literal query and plausible implicit need, then uses that estimate to direct per-query probing, tool selection, and optional heads-up alerts.

\textbf{Contributions.}
\textbf{(i)} We introduce the \texttt{IntentFrame} as a \emph{pre-tool control variable}: before any tool fires, the agent infers the user's implicit need and emits a scalar gap score that determines the probe budget and shortlists candidate tools. 
\textbf{(ii)} On a 100-query four-scene implicit-intent benchmark, gap-routed probing improves implicit-need coverage over ReAct-style NoIntent ($\Delta{=}{+}0.07$, $p{<}10^{-6}$; three of four scenes significant), reproducing on a 25-query pilot, a second backbone, and under disjoint-example ablation.
\textbf{(iii)} We characterise the regime boundary: an access--cost Pareto point on factual grounding, not a universal accuracy win (\S\ref{sec:rq2}; Limitations).
\textbf{(iv)} We release the \textsc{AURATown} simulator, all 100 implicit-intent queries with subcategory labels (inter-annotator agreement $\kappa{=}0.61$, measured on the 25-query pilot subset), and per-seed run records.

\section{Related Work}
\label{sec:related}

\paragraph{Agent architectures and tool use.}
ReAct, Reflexion, Toolformer, AgentBench, ToolLLM, T-Eval~\citep{yao2023react, shinn2023reflexion, schick2024toolformer, liu2023agentbench, qin2024toolllm, chen2024teval} treat tool use as reactive: tools fire \emph{during} reasoning. \sys{}'s Explore runs \emph{before} reasoning with a bounded budget routed by the inferred literal-vs.-implicit gap. Plan-and-Solve~\citep{wang2023plansolve} plans tool calls from the surface query without that gap routing.

\paragraph{Multi-agent simulation and memory.}
Generative Agents~\citep{park2023generative} inject environmental observations into LLM prompts; \sys{} adds (i) per-query access \emph{control} by an LLM-produced \texttt{IntentFrame}, vs.\ passive top-$k$ retrieval; (ii) a public/private state split that Park's all-public state lacks. SOTOPIA~\citep{zhou2024sotopia} provides the 7-dimension framework we adopt; broader multi-agent benchmarks~\citep{zhou2024sotopia, zhu2025marble, xu2024magic, vezhnevets2023concordia, chen2024agentverse, li2023camel, hong2024metagpt} treat the environment as serving agents on request. Memory frameworks~\citep{packer2023memgpt, zhong2024memorybank, lewis2020retrieval, gao2023retrieval, bmam2026prior} target long-term storage; \sys{}'s memory is populated by proactive probing, not passive accumulation. Proactive context agents~\citep{yang2025contextagent, proagent2025sensory, probe2025proactive, wang2026proagentbench} (Appendix~\ref{app:proactive-comparison}) and interactivity-native systems~\citep{tmlab2026interaction} target \emph{when to assist}; \sys{} is orthogonal, targeting \emph{what private-state context to surface} for an already-received query.

\paragraph{Pragmatic and conditional QA.}
Text-only QA meets the same motivation: users omit context they assume the model shares. \textsc{CondAmbigQA}~\citep{li2025condambigqa} resolves this by condition-aware reasoning over text; in \sys{} the implicit need lives in another agent's hidden private state, and the resolution is a budgeted probe over an environment registry rather than a textual condition rewrite.

\paragraph{Intent, plan recognition, and information value.}
The IntentFrame revisits three older ideas. \emph{Plan recognition}~\citep{kautz1986plan, charniak1993bayesian, carberry2001plan, baker2009inverse} infers an agent's goal from observed actions; our inferrer infers the \emph{user's} unstated need from a query and a scene, and acts on it by probing. In pragmatics, the question-under-discussion framework~\citep{roberts2012qud} and Gricean implicature~\citep{grice1975logic} formalise how a literal utterance underdetermines what is being asked; rational-speech-act models~\citep{frank2012pragmatic} make that inference probabilistic; \sys{} collapses it to a scalar control input. POMDP dialogue managers~\citep{williams2007pomdp, young2013pomdp} choose information-gathering actions under belief uncertainty, and value-of-information analysis~\citep{howard1966information}, applied to clarification questions by \citet{rao2018clarification}, prices each such action. Our gap-to-budget map is a hand-tuned stand-in for that computation, and the inverted calibration we report (\S\ref{sec:rq-tom}) is the symptom a learned value-of-information estimate would address.

\paragraph{Belief-state evaluations.}
Theory of mind in LLMs is contested: \citet{kosinski2023llmtom} argues spontaneous emergence; \citet{sap2022neuraltom, ullman2023largelanguagetom} show failure on trivially-altered Sally--Anne tasks; \citet{sclar2023symbolictom} build an explicit symbolic tracker. \sys{} is \emph{not} a ToM benchmark: of our five implicit-intent subcategories, only \emph{second\_order} probes belief tracking (stale-belief templates adapted from \citet{ullman2023largelanguagetom}); the other four (\emph{availability, mood, appropriateness, latent\_goal}) target surfacing of non-belief private state through a budgeted controller. We position \sys{} as evidence about \emph{when intent-directed probing helps surface hidden state}; belief queries appear in our 5-subcategory taxonomy as one substrate (\emph{second\_order}). Transfer to external belief benchmarks does not hold automatically: Appendix~\ref{sec:appendix-fantom} reports a FANToM transfer null.
\section{The \sys{} Framework}
\label{sec:method}

\begin{figure*}[!t]
\centering
\includegraphics[width=0.68\textwidth]{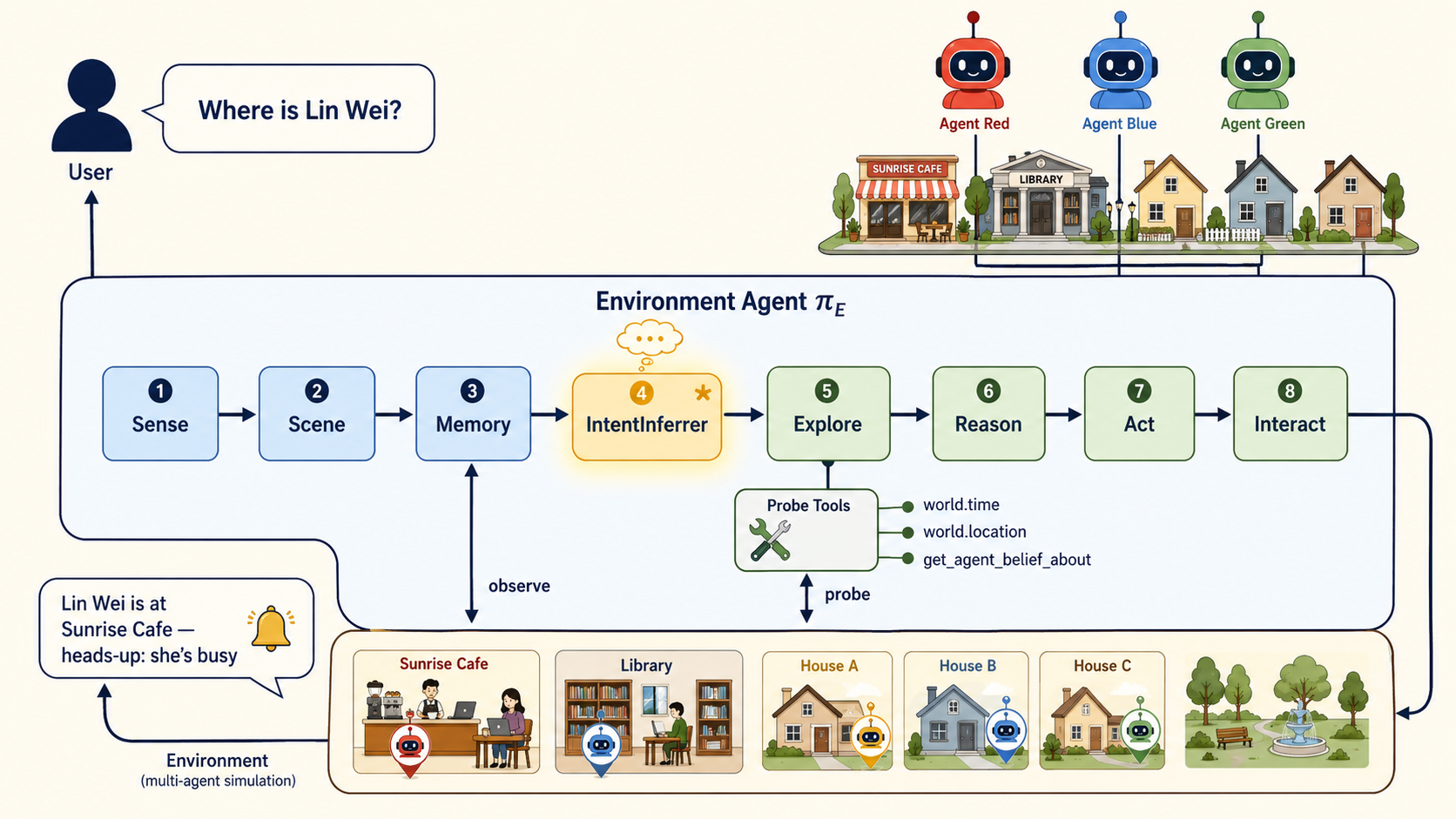}
\caption{\sys{} pipeline. Sense receives the user query; Scene and Memory assemble a deterministic context preview. The \textbf{IntentInferrer} ($\star$) outputs an \texttt{IntentFrame} whose scalar \texttt{gap} field routes probe budget and tool selection in Explore (Section~\ref{sec:method-intent}). Reason plans over the enriched context; Act and Interact emit the response.}
\label{fig:architecture}
\end{figure*}

\sys{} factors a situated agent into two phases (Fig.~\ref{fig:architecture}): \emph{deterministic context assembly} (Sense $\to$ Scene $\to$ Memory) followed by \emph{LLM-controlled reasoning} (IntentInferrer $\to$ Explore $\to$ Reason $\to$ Act $\to$ Interact). The deterministic phase establishes a passive context preview using the same passive-perception pattern as Generative Agents~\citep{park2023generative} and is not the contribution. The contribution is the \textsc{IntentInferrer}, which produces an \texttt{IntentFrame} whose scalar gap field controls per-query probe budget and tool selection in Explore (Section~\ref{sec:method-intent}); Reason, Act, and Interact then plan and emit the response, with an optional heads-up alert when the inferrer's gap value crosses a threshold. The remainder of this section formalises the problem (Section~\ref{sec:formulation}), describes the bounded-probing loop (Section~\ref{sec:probing}), and details the \texttt{IntentFrame} stage (Section~\ref{sec:method-intent}).

\subsection{Problem Formulation and Architecture}
\label{sec:formulation}
\label{sec:architecture}

We consider a situated multi-agent system where $N$ agents operate in a shared environment $\mathcal{E}$ over discrete time steps. A human user $h$ may interact with any agent via natural-language queries $q$. We seek to maximise \emph{environmental grounding} of agent actions and user-facing responses:
\begin{equation*}
\max_{\pi_E}\mathbb{E}\!\left[\sum_t G(\alpha_i^t, \mathcal{E}^t) + \sum_q R(r_q, \mathcal{E}^t_q)\right],
\end{equation*}
where $G$ measures action-state consistency, $R$ measures response factual accuracy, and $\pi_E$ is the \envagent{}'s probing policy.

The Explore stage's useful range is bounded by the residual uncertainty after passive perception: when Sense/Scene/Memory cover the facts the query needs, extra probes add little; when the query depends on a remote agent's private state, one targeted probe can change the answer.

\subsection{Bounded Proactive Probing}
\label{sec:probing}

The \textbf{Explore} stage implements bounded environment probing. Unlike ReAct~\citep{yao2023react} where tool calls are interleaved \emph{during} reasoning, our probing occurs \emph{before} answer generation, so the final response is conditioned on a compact probe trace. The probe loop iterates up to a budget $B$ steps; at each step an LLM planner $\phi$ examines the current context and the tool registry $\mathcal{T}{=}\{t_1,\dots,t_K\}$ and either issues a tool call (whose result is appended to a probe trace and used to update the context) or signals \texttt{stop}; the trace is summarised at the end and consumed by Reason. Pseudocode in Algorithm~\ref{alg:probe} (Appendix~\ref{app:algorithm}).

\paragraph{Design.} The loop is goal-directed (LLM-driven tool selection given current context, not exhaustively), bounded (step budget $B$ caps API cost), composable (each tool result updates context for the next decision, enabling multi-hop gathering), and separable (probing runs independently from reasoning, so results can be cached or skipped). The \textsc{AURATown} instantiation registers eight base environment tools (Table~\ref{tab:tools}) for the factual-grounding benchmark, and a separate five-tool scripted registry of agent-state probes for the implicit-intent benchmark (Section~\ref{sec:rq-tom}); both register under a pattern-based allow/deny policy.

\subsection{Intent Inference: Modeling the User's Implicit Need}
\label{sec:method-intent}

Proactive probing decides \emph{what to ask the environment}; intent
inference estimates \emph{what useful answer the user may need}.
\sys{} factors the second decision out as an explicit LLM-mediated
stage that sits between Scene/Memory and the Explore budget selector.

Given the user's surface query $q$, the preview scene $\mathcal{S}$,
and recent memories $M$, the \texttt{IntentInferrer} produces an
\texttt{IntentFrame}:
\begin{align*}
\texttt{IntentFrame} =\, & (\ell,\; I,\; g \in [0,1],\; P \subseteq \mathcal{T},\\
& \;a \in \{0,1\},\; c \in [0,1],\; r)
\end{align*}
where $\ell$ is the literal-need restatement, $I$ is the list of
plausible implicit needs, $g$ is the \emph{gap}
between literal and implicit (0 if the literal answer suffices, 1 if
the user's real need is orthogonal), $P$ is a set of recommended probe
tools drawn from the registry $\mathcal{T}$, $a$ is an alert flag, $c$
is the inferrer's self-reported confidence, and $r$ a rationale.

The gap $g$ is the control input: downstream stages use a deterministic
map $B(g) = 0, 1, 2, 3, 5$ for $g \in [0,0.2),[0.2,0.4),[0.4,0.6),[0.6,0.8),[0.8,1]$
respectively, truncated by the global \texttt{explore\_max\_steps}
budget. Importantly, $B(g)$ is a \emph{ceiling}, not a target: the
downstream Explore loop receives both the probe budget and the
\texttt{recommended\_probes} hint, and we observe in
Section~\ref{sec:adaptive-budget} that the LLM typically stops short
of the ceiling when one well-targeted probe has already returned
actionable information. The two signals are separable, and the
component ablation in Section~\ref{sec:rq-tom} shows they are not
equally load-bearing: the gap-derived ceiling carries the gain, while
the recommended-probes shortlist is dispensable, and on our registry
slightly harmful, once budget routing is in place.

\paragraph{Worked example.}
For the query ``where is Lin Wei?'' at 14:30 with Lin Wei present in the cafe scene, the inferrer outputs $\ell{=}$``locate Lin Wei'', $I{=}$\{``is she free to chat?'', ``is she taking a break?''\}, $g{=}0.6$, $P{=}$\{\texttt{get\_agent\_private\_state}, \texttt{get\_agent\_plan}\}, $a{=}1$, $c{=}0.7$. The $g{=}0.6$ maps to a budget ceiling of $B{=}3$, but Explore stops after two probes when \texttt{get\_agent\_private\_state(``Lin Wei'')} returns \texttt{availability=busy}, which suffices to answer the implicit need. 

\paragraph{Two backends and control boundary.}
A heuristic backend (deterministic surface-cue matching, for tests and offline execution) and an LLM backend (structured-JSON output, calibrated with a four-tier gap rubric and clean benchmark-disjoint few-shot exemplars; recommended-probes whitelisted against the live tool registry) are both provided. The architectural pipeline up to this point (Sense--Scene--Memory) is deterministic; the \texttt{IntentFrame} is the first stage at which the LLM affects control flow (probe budget, tool priority, alert flag). Section~\ref{sec:adaptive-budget} quantifies the resulting per-query adaptation, and Appendix~\ref{sec:appendix-prompt-ablation} shows that the examples act as gap calibration rather than answer templates. The heuristic backend is provided for offline tests and air-gapped deployments; all reported results use the LLM backend.

\paragraph{Memory architecture.}
\label{sec:memory}
\sys{}'s memory scores each item $m$ against query $q$ at time $t$ by a weighted combination of recency, importance, and lexical similarity (Eq.~\ref{eq:memory_score}, Appendix~\ref{sec:appendix-memory}, with weights $0.3{:}0.4{:}0.3$ and decay $0.01$). Memory types follow~\citet{tulving1972episodic} (observation, conversation, reflection, plan), with reflections triggered every $\theta{=}10$ observations. Keyword-based similarity is a deliberate reproducibility trade-off; embedding retrieval is a natural extension. A three-stage enrichment protocol routes each query through context gathering, single-step probe verification, and enriched generation (Appendix~\ref{sec:appendix-memory}).

\paragraph{Relationship to long-context approaches.}
A natural alternative is to pack the full environment state into the prompt and rely on the LLM's long-context attention. \sys{}'s selective probing pays one extra IntentInferrer call but issues 0--3 targeted probes whose count tracks the inferred gap, rather than the world size. We do not run a head-to-head against a stuff-everything baseline (such a baseline would also need to decide which slice of memory to include); the Static-Context baseline (\S\ref{sec:rq2}) (a packed scene snapshot) is the closest in-suite analogue and reaches FA $0.450$ vs.\ \sys{} Full's $0.640$. A retrieval-packing baseline that selects a larger private-state slice under a \emph{matched token budget} is the comparison that would directly test selective probing against context packing; we name it as open future work rather than approximate it here. Appendix~\ref{app:longcontext} sketches the token-cost and prompt-content differences.

\section{AURATown: A Multi-Agent Social Simulation}
\label{sec:auratown}

\textsc{AURATown} is a 60$\times$60 grid-based social simulation with 5 named agents and 20 named locations over a 6:00--23:00 day (full setup, map, and a tick-18:00 mechanism snapshot in Appendix~\ref{app:agents}). Two decisions depart from \citet{park2023generative}'s 25-agent Smallville. \emph{(i) Scale.} 5 agents yields tractable per-query ground truth; we test a per-agent mechanism, not multi-agent emergence. \emph{(ii) Public/private state split.} \texttt{location} and \texttt{action} are visible in the scene snapshot, but \texttt{availability}, \texttt{emotional\_state}, \texttt{unspoken\_goal}, and \texttt{beliefs\_about\_others} are hidden and only retrievable via probe tools. Each agent's private state updates deterministically each tick via a seven-rule priority table (Appendix~\ref{app:private-state-rules}): e.g., an agent at a loaded workplace becomes busy and tired-focused; one at an empty workplace becomes available and lonely. Beliefs about other agents refresh only on co-location, producing the staleness that \emph{second\_order} queries probe. The released codebase additionally includes a chunk-based procedural world used by the demo deployment (\texttt{demo/town/chunks.py}); experiments here use only the fixed 60$\times$60 / 5-agent / 20-location subset.

\section{Experiments}
\label{sec:experiments}

We evaluate \sys{} along three axes: environment access as a boundary condition (\S\ref{sec:rq2}), intent-directed probing (\S\ref{sec:rq-tom}), and adaptive probe allocation. Primary experiments use \texttt{gpt-4o-mini} as both agent backbone and LLM-as-judge (full hyperparameters in Appendix~\ref{app:implementation}; cross-backbone robustness on \texttt{claude-haiku-4-5}, \texttt{qwen-plus}, \texttt{gemini-2.5-flash} in Appendix~\ref{sec:appendix-cross-backbone}). The same-family judge--policy setup limits independence; we address this with a strict precision rescore in Appendix~\ref{sec:appendix-strict-rescore}. Baselines, metrics, and additional diagnostic checks (routine grounding, component ablation, SOTOPIA, human eval, budget sweep) are in Appendices~\ref{sec:appendix-setup}--\ref{sec:appendix-minor-rqs}.

\paragraph{Regime characterisation.}
A scope condition governs everything that follows and should be read as a \emph{precondition} of the method, not a footnote: the IntentFrame controller assumes an environment whose task-relevant hidden state is \emph{tool-gated}: absent from the prompt context, but retrievable through a registered probe. \sys{} Intent leads the implicit-need regime (\S\ref{sec:rq-tom}), which satisfies this condition by construction; on factual grounding (\S\ref{sec:rq2}), gap-routed probing is an access-cost Pareto point rather than an accuracy winner. Where the condition fails the mechanism adds nothing: no lift when private state is already in context (LoCoMo) or structurally inaccessible (FANToM, GAIA; Appendices~\ref{sec:appendix-fantom}--\ref{sec:appendix-gaia}).

\paragraph{Routine grounding: metric saturates.}
Under 100 simulation steps per condition $\times$ 3 seeds, all five architectures (Vanilla, Static Context, ReAct, \sys{} No-Probe, \sys{} Full) fall within $0.024$ absolute GA spread, every paired $t$-test vs.\ vanilla gives $p > 0.5$. The metric is saturated: memory-utilisation $\approx 1.0$ for every method, most actions are trivially grounded (``sleeping at home at 6\,AM''). This is evidence about the metric, not the mechanism (Appendix~\ref{sec:appendix-rq1}).

\subsection{Factual Grounding: Environment Access as a Boundary Condition}
\label{sec:rq2}

This subsection tests whether gap-routed probing is merely a generic accuracy booster (it is not). We collect 50 user queries about the environment spanning 5 categories (spatial, social, temporal, memory, planning). Responses are scored against ground-truth environment state by a \texttt{gpt-4o-mini} LLM-as-judge (same family as the agent backbone --- see Appendix~\ref{sec:appendix-strict-rescore} for a strict precision rescore that reduces sensitivity to judge softness). We repeat across three random seeds $\{42, 123, 456\}$ and report mean $\pm$ std, together with paired $t$-tests against two reference conditions: the vanilla LLM (to measure total architectural effect) and \sys{} (No Probe) (to isolate the marginal contribution of proactive probing on top of the Perceive/Scene/Memory pipeline).

\begin{table*}[!tp]
\centering
\caption{Factual Accuracy (FA) on environment-grounded queries ($N{=}50$, 3 seeds). \textbf{Bold} marks within-column best by criterion: Fixed-Probe / Plan-and-Solve on raw FA; GapRouted infers a per-query gap and executes only the recommended probes;\sys{} GapRouted ($\star$, Pareto frontier in Fig.~\ref{fig:rq2-pareto}) on access cost (Probes); ReAct on contradicted-claim rate. Contrad.\,\% is diagnostic; \sys{} makes no hallucination claim in this regime. Strict precision rescore in Appendix~\ref{sec:appendix-strict-rescore}. $^{\dagger}$ReAct calls tools via OpenAI function-calling, bypassing the AURATown probe registry; its probe count is not directly comparable.}
\label{tab:hallucination}
\small
\begin{tabular}{lcccccc}
\toprule
\textbf{Method} & \textbf{FA} & \textbf{Probes} & \textbf{Contrad.\,\%} & \textbf{Lat.\,(s)} & \textbf{$p_{\text{vs.\ Van}}$} & \textbf{$p_{\text{vs.\ Full}}$} \\
\midrule
Vanilla LLM            & $0.070 \pm 0.010$ & $0.00$  & $93.3$ & $\phantom{0}2.3$ & --- & $<10^{-4}$ \\
Static Context         & $0.450 \pm 0.043$ & $0.00$  & $70.0$ & $\phantom{0}1.8$ & $0.032$ & $<10^{-3}$ \\
ReAct Agent            & $0.550 \pm 0.037$ & ---$^{\dagger}$ & $\mathbf{51.3}$ & $\phantom{0}6.0$ & $0.100$ & $0.033$ \\
Reflexion              & $0.174 \pm 0.049$ & $2.15$  & $55.9$ & $20.6$           & --- & $<10^{-4}$ \\
Plan-and-Solve         & $0.764 \pm 0.020$ & $4.64$  & $58.7$ & $\phantom{0}5.2$ & --- & $0.004$ \\
Fixed-Probe            & $0.766 \pm 0.021$ & $8.00$  & $56.0$ & $\phantom{0}3.2$ & $<10^{-4}$ & $0.006$ \\
\sys{} (No Probe)      & $0.603 \pm 0.028$ & $0.00$  & $75.3$ & $\phantom{0}2.2$ & $0.0005$ & $0.060$ \\
\sys{} (Full, $B{=}2$) & $0.640 \pm 0.021$ & $2.00$  & $66.7$ & $\phantom{0}3.9$ & $0.0010$ & --- \\
\sys{} (GapRouted)$^\star$ & $0.696 \pm 0.026$ & $\mathbf{1.40}$ & $75.3$ & $\phantom{0}4.3$ & $<10^{-4}$ & $0.098$ \\
\bottomrule
\end{tabular}
\end{table*}

\begin{figure*}[!t]
\centering
\includegraphics[width=0.92\textwidth]{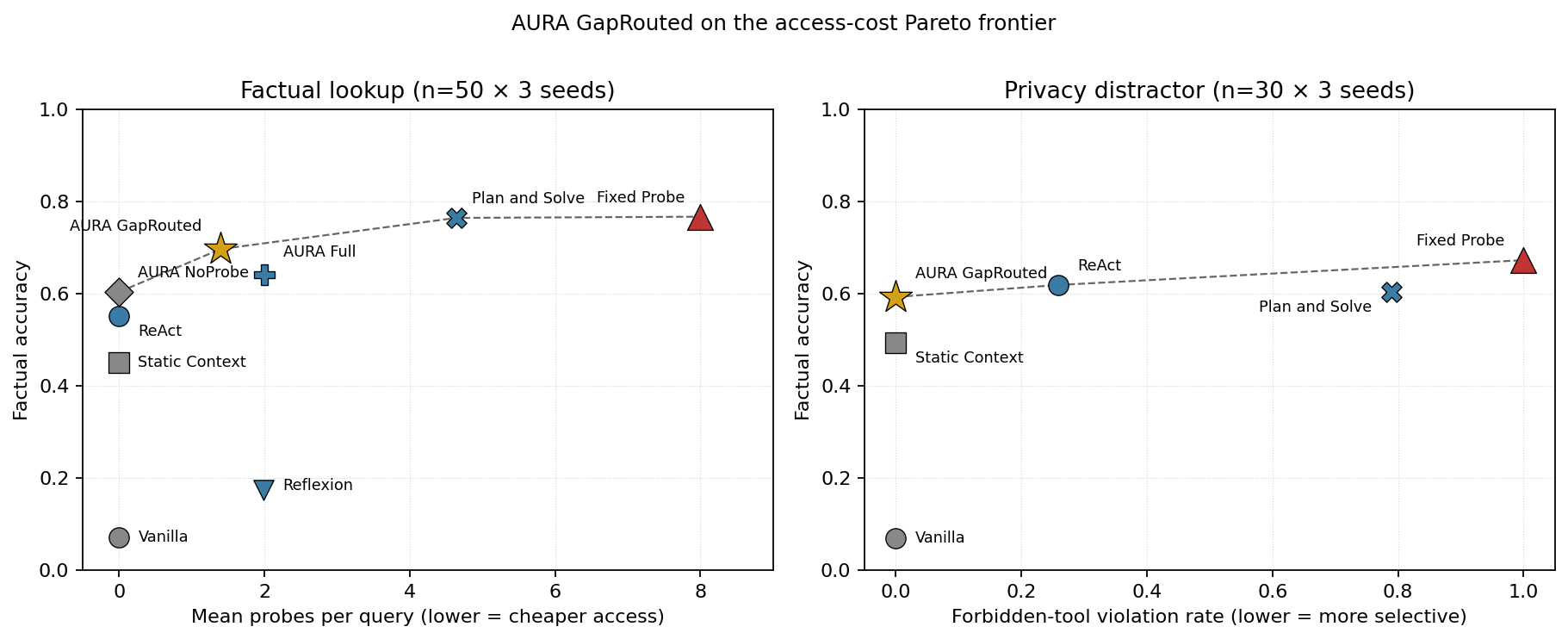}
\caption{Factual-grounding access Pareto. Left: factual lookup ($N{=}50 \times 3$ seeds) --- factual accuracy vs.\ mean probes per query. \sys{} GapRouted ($\star$) sits on the frontier with $1.40$ probes; Fixed-Probe wins raw FA at $8\times$ the access cost. Right: privacy-sensitive distractor slice ($N{=}30 \times 3$ seeds) --- factual accuracy vs.\ forbidden-tool violation rate. \sys{} GapRouted holds the Pareto vertex with $0\%$ violations while approaching Fixed-Probe ($100\%$) and Plan-and-Solve ($78.9\%$) in FA.}
\label{fig:rq2-pareto}
\end{figure*}

\paragraph{Scope evidence: factual lookup is an accuracy--access tradeoff regime.}
The 50 factual queries ask for environment facts --- ``Where is X?'', ``What time is it?'' --- so saturated access is a strong control. Fixed-Probe invokes all eight environment tools on every query and reaches $0.766$ FA, statistically above GapRouted ($0.696$; paired $\Delta{=}{+}0.070$, $p{=}0.031$). Plan-and-Solve is similar in raw FA ($0.764$; paired vs.\ GapRouted $\Delta{=}{+}0.068$, $p{=}0.056$). The gap-routed controller therefore is \emph{not} the raw-accuracy winner on factual lookup. Its contribution is an access Pareto point (Figure~\ref{fig:rq2-pareto}): $1.40$ probes/query vs.\ Fixed-Probe's $8.00$ ($\mathbf{82\%}$ fewer; $p{=}4.9\times10^{-52}$), and disclosure $0.92$ vs.\ $5.00$ ($p{=}6.7\times10^{-38}$). 
\emph{Disclosure} is a per-query sum, over the tools actually invoked, of a fixed sensitivity weight: $0$ for public-observable tools (\texttt{world.time}, \texttt{world.location}, \texttt{world.\allowbreak{}nearby\_agents}), $1$ for moderate ones (\texttt{world.\allowbreak{}agents\_summary}, \texttt{world.\allowbreak{}events\_recent}, \texttt{agent.plan}), and $2$ for \texttt{memory.recent}; unlike raw probe count, it weights \emph{what kind} of state each call exposes.
On a 30-query privacy-sensitive factual slice (Appendix~\ref{sec:appendix-privacy}), GapRouted ties Plan-and-Solve and ReAct in FA ($p{=}0.86$ and $0.65$) while reducing forbidden-tool violations to $\mathbf{0\%}$ (Plan-and-Solve $78.9\%$, ReAct $25.6\%$, Fixed-Probe $100\%$). Per-query wall-clock latency tells a more cautious story: GapRouted pays the IntentInferrer's extra LLM round trip and is slower at the median than Fixed-Probe ($4.08$ vs.\ $2.37$\,s; Appendix~\ref{sec:appendix-cost-latency}), so the cost-of-selectivity claim holds on probe count and disclosure rather than on latency. The mechanism's primary accuracy contribution remains the implicit-need setting in Section~\ref{sec:rq-tom}; the factual-grounding regime bounds the cost of selective access.

\paragraph{Reflexion failure mode.}
Reflexion (Table~\ref{tab:hallucination}) collapses to $0.174$ FA because the reflection step withdraws probe-supported claims (Appendix~\ref{sec:appendix-rq2-cat}); the Full-vs-No-Probe contrast is near-null at gap${\approx}0$ ($p{=}0.060$).

\paragraph{Privacy as a design benefit.}
Gap routing's zero forbidden-tool violation rate is not a post-hoc observation: low-gap factual queries receive $B{=}0$ budget and skip the probe registry entirely, so they structurally cannot invoke a forbidden tool. This makes the access--cost Pareto a privacy Pareto as well.

\paragraph{Strict-precision robustness.}
A strict-precision rescore (Appendix~\ref{sec:appendix-strict-rescore}) confirms the access-cost picture: per-category gains concentrate on temporal and memory; the overall architectural margin above ReAct narrows to non-significance ($\Delta{=}{+}0.090$, $p{=}0.058$). The IntentFrame's primary accuracy contribution is the implicit-need regime (\S\ref{sec:rq-tom}).

\subsection{Implicit-Need Surfacing via Intent Inference}
\label{sec:rq-tom}

This subsection tests whether intent inference itself surfaces the user's implicit information need on queries specifically designed to hide that need behind a literal surface form.

\paragraph{Query set.}
We construct 25 primary implicit-intent queries stratified across five subcategories that vary the relationship between surface form and implicit need: \emph{availability} (e.g., ``where is Lin Wei?'' $\to$ ``is she free?''), \emph{mood}, \emph{appropriateness}, \emph{latent\_goal}, and \emph{second\_order} (representative queries for the remaining four and the second-order belief-vs-truth judging rule in Appendix~\ref{sec:appendix-rq-intent-v2}). The scene exposes only public state (location, action); private-state fields (availability, emotional state, unspoken goal, beliefs about others) are only retrievable via probe tools.

\paragraph{Conditions.}
Four answer strategies share the same backbone (gpt-4o-mini, temperature 0.1) and a 5-tool agent-state registry (Appendix~\ref{app:tools}; disjoint from the factual-grounding registry): \textbf{Literal} (scene only, no tools); \textbf{NoIntent} (ReAct-style, up to $B{=}3$ tool calls); \textbf{Plan-and-Solve}~\citep{wang2023plansolve} (plan/execute/synthesise, budget 3); \textbf{Intent} (IntentInferrer $\to$ gap-derived probe ceiling 0--3 $\to$ directed tool loop, with heads-up prefix on high-gap queries).
Each response is scored by a \texttt{gpt-4o-mini} judge on two dimensions in $[0,1]$: \texttt{literal\_score} and \texttt{implicit\_score} (does it surface the implicit need?). We run 3 seeds (42, 123, 456) over the 25 queries and report query-level paired $t$-tests as primary, with seed-level paired tests and \texttt{query\_id} cluster bootstraps as secondary checks (Appendix~\ref{sec:appendix-rq-intent-v2}).

\begin{table}[!t]
\centering
\caption{Implicit-intent comparison ($N{=}25$, 3 seeds; scores in $[0,1]$). \textbf{Bold} marks the best deployable system per column; Fixed-private and Oracle-intent are diagnostic upper bounds (require unconditional private-state access or gold-tool knowledge) and not bolded. Tool-using rows share the same 5-tool registry. Prompt-example and component ablations in Appendices~\ref{sec:appendix-prompt-ablation},~\ref{sec:appendix-minor-rqs}.}
\label{tab:rq-tom-overall}
\footnotesize
\resizebox{\columnwidth}{!}{%
\begin{tabular}{lcccc}
\toprule
\textbf{Condition} & \textbf{Lit.} & \textbf{Implicit} & \textbf{Probes} & \textbf{Lat.} \\
\midrule
\multicolumn{5}{l}{\emph{Deployable systems}} \\
Literal           & $0.659$           & $0.216$           & $0.00$ & $1.3$ \\
NoIntent          & $0.947$           & $0.640$           & $1.04$ & $2.2$ \\
Plan-and-Solve    & $0.811$           & $0.520$           & $1.37$ & $11.1$ \\
\sys{} Intent     & $\mathbf{0.957}$  & $\mathbf{0.803}$  & $1.48$ & $13.8$ \\
\midrule
\multicolumn{5}{l}{\emph{Diagnostic upper bounds (not deployable)}} \\
Fixed-private     & $0.987$           & $0.851$           & $1.40$ & $2.4$ \\
Oracle-intent     & $0.997$           & $0.861$           & $2.32$ & $2.7$ \\
\bottomrule
\end{tabular}
}
\end{table}

\begin{table}[!t]
\centering
\caption{100-query four-scene implicit-intent benchmark (3 seeds, \texttt{gpt-4o-mini}). Implicit-need coverage scores in $[0,1]$. \textbf{Bold} marks best per column. $p$ values are paired tests on query-seed cells.}
\label{tab:rq-intent-v2-main}
\footnotesize
\setlength{\tabcolsep}{3pt}
\resizebox{\columnwidth}{!}{%
\begin{tabular}{lccccc}
\toprule
 & \textbf{Overall} & \textbf{A: cafe} & \textbf{B: library} & \textbf{C: garden} & \textbf{D: night} \\
\midrule
Literal          & $0.223$ & $0.200$ & $0.264$ & $0.237$ & $0.192$ \\
NoIntent         & $0.733$ & $0.709$ & $0.688$ & $0.731$ & $0.803$ \\
\sys{} Intent    & $\mathbf{0.804}$ & $\mathbf{0.800}$ & $\mathbf{0.779}$ & $\mathbf{0.811}$ & $\mathbf{0.827}$ \\
\midrule
$\Delta$ (Intent--NoIntent) & $+0.071$ & $+0.091$ & $+0.091$ & $+0.080$ & $+0.024$ \\
$p$                      & $1.0{\times}10^{-6}$ & $0.004$ & $1.4{\times}10^{-4}$ & $0.015$ & $0.349$ \\
\bottomrule
\end{tabular}%
}
\end{table}
\paragraph{Overall.}
On the 100-query four-scene benchmark (Table~\ref{tab:rq-intent-v2-main}), \sys{}-Intent reaches $\mathbf{0.804}$ implicit-need coverage versus $0.733$ for NoIntent (ReAct-style) and $0.223$ for Literal: paired $\boldsymbol{\Delta{=}{+}0.071}$, $\boldsymbol{p{=}1.0\times10^{-6}}$. Three of four scenes are individually significant; night scene D ties because public state already telegraphs availability. A 25-query pilot (Scene~A of the four-scene set, Table~\ref{tab:rq-tom-overall}) confirms the direction at higher absolute gain ($\Delta{=}{+}0.16$, $p{=}0.006$) and additionally tests Plan-and-Solve ($\Delta{=}{+}0.28$, $p{=}8.3\times10^{-5}$) and two diagnostic upper bounds (fixed-private $0.851$, oracle-intent $0.861$; underpowered to separate from \sys{} at $N{=}25$).

\paragraph{Per-subcategory structure.}
The subcategory breakdown shows where intent inference helps. \emph{Availability} gains most ($\Delta{=}{+}0.29$, $p{=}2.7\times10^{-11}$): ``where is X?'' fully masks the implicit need, so the gap score routes a probe that would not otherwise fire. \emph{Appropriateness} ($+0.11$) and \emph{mood} ($+0.07$) follow at smaller magnitudes (both $p{<}2\times10^{-3}$). \emph{Second\_order} ties ($-0.02$, NS): ``does X think Y\ldots'' already cues belief-state probing. \emph{Latent\_goal} shows a deficit ($-0.09$, $p{=}6.2\times10^{-4}$). The gap score adds value precisely when the surface form is maximally decoupled from the implicit need. A worked failure case for \emph{latent\_goal}, where the shortlist spends a full ceiling on one agent and misses the intent, is in Appendix~\ref{sec:appendix-rq-intent-v2}.

\paragraph{Few-shot calibration is a precondition of the method.}
A three-way prompt ablation (Appendix~\ref{sec:appendix-prompt-ablation}) shows the gain does not come from memorising examples, but does depend on their presence. With clean benchmark-disjoint exemplars the Intent-vs-NoIntent contrast is $\Delta{=}0.163$ ($p{=}0.006$); removing them entirely reduces it to $\Delta{=}0.035$ ($p{=}0.44$). Replacing them with leaked examples changes little, which rules out answer-template memorisation. What the exemplars supply is calibration: without them the mean inferred gap falls and half as many queries clear the probe threshold. Any prompted control variable needs some such signal, and ours is a small, disclosed, non-answer-specific one; the method should not be expected to work without it.

\paragraph{Cross-backbone.}
The gain reproduces on \texttt{claude-haiku-4.5} ($\Delta{=}{+}0.086$, $p{=}3.7\times10^{-3}$) and \texttt{qwen-plus} ($\Delta{=}{+}0.25$, CI $[+0.14,+0.37]$); per-backbone detail and a Gemini JSON-parse failure in Appendix~\ref{sec:appendix-cross-backbone}.

\paragraph{Backend ablation: gap inference is load-bearing.}
Replacing the \texttt{LLMIntentInferrer} with a deterministic \texttt{HeuristicIntentInferrer} (rule-based gap, identical plumbing) drops implicit score from $\mathbf{0.803}$ to $0.368$ (Appendix~\ref{sec:appendix-heuristic}), most on lexically-decoupled subcategories: the lift attaches to LLM-mediated gap inference, not the scaffolding.

\paragraph{Human evaluation: output quality vs.\ NoIntent.}
Eight raters scored 50 paired (\sys{} vs.\ Vanilla) scenarios on four dimensions; \sys{} rates higher on all four, with $74\%$ of cells reaching ${\geq}6/8$ consensus for \sys{} and none for Vanilla (Appendix~\ref{sec:rq5}). That establishes only a preference over the weakest baseline, so we ran a second arm against NoIntent on the implicit-intent items, with the scene's ground-truth state shown so that accuracy was checkable (Appendix~\ref{sec:rq5-nointent}). It \textbf{does not} establish a human-perceptible advantage: three dimensions favour \sys{} without reaching significance, and \sys{} rates significantly lower on factual accuracy ($p{=}0.008$), where no rater preferred it. Two things bear on reading that: the arm rated the full \texttt{IntentFrame} variant, the one budget-only routing dominates, and \sys{}'s answers run nearly twice as long, which raters penalised. Coverage of a stated hidden need and holistic response quality are distinct properties, and \sys{} claims only the first.

\paragraph{What the coverage gain does not buy.}
Two results should be read together. \sys{} surfaces more of the user's implicit need, and it also carries a higher contradicted-claim rate than ReAct on factual lookup (Table~\ref{tab:hallucination}) and rates lower on human-judged factual accuracy. We take the straightforward reading rather than the flattering one: the system says more, and a larger share of what it says is unsupported by what its probes actually returned. The coverage metric scores whether the stated hidden need was addressed; it does not penalise the elaboration surrounding that answer, and our heads-up prefix actively encourages such elaboration. Coverage and grounding are therefore not two views of one improvement, and a deployment that cares about the second should add an output constraint we do not impose: assert only what a probe returned.

\paragraph{Adaptive budget: ceiling, not target.}
Mean probes per query range $0.80$--$2.20$ under a fixed ceiling of 3, and gap correlates only $\mathbf{r{=}0.19}$ with realised probe count: the gap routes a ceiling, not a target, so the agent spends less than a fixed-budget system on most queries (Appendix~\ref{sec:adaptive-budget}).

\begin{table}[!t]
\centering
\caption{Component ablation, 100-query benchmark ($n{=}300$ query--seed cells). \emph{Budget} = gap-derived ceiling; \emph{Hints} = LLM-recommended tools $P$; without hints the loop uses Fixed-Probe's deterministic tool order. Paired $t$-tests vs.\ NoIntent.}
\label{tab:component-ablation}
\footnotesize
\setlength{\tabcolsep}{4pt}
\resizebox{\columnwidth}{!}{%
\begin{tabular}{lcccc}
\toprule
\textbf{Condition} & \textbf{Implicit} & \textbf{Probes} & $\Delta$ \textbf{vs.\ NoIntent} & $p$ \\
\midrule
NoIntent (neither)      & $0.733$          & $1.11$ & ---      & --- \\
Budget only             & $\mathbf{0.827}$ & $1.20$ & $+0.095$ & $6.9{\times}10^{-11}$ \\
Hints only              & $0.813$          & $1.74$ & $+0.080$ & $5.6{\times}10^{-8}$ \\
Both (\sys{})           & $0.804$          & $1.75$ & $+0.071$ & $6.4{\times}10^{-7}$ \\
\bottomrule
\end{tabular}%
}
\end{table}

\paragraph{Component ablation: the budget ceiling, not the probe hints, carries the gain.}
The \texttt{IntentFrame} feeds two signals downstream: the gap-derived budget ceiling $B(g)$ and the recommended-probe shortlist $P$. Isolating them (Table~\ref{tab:component-ablation}) shows the ceiling alone, paired with Fixed-Probe's deterministic tool order, is the strongest configuration; the full system is significantly worse, and consistently so across four of five subcategories. This is not an artifact of unstable inference: the two conditions infer near-identical gap values, yet agree on only a fifth of their realised tool sequences. The shortlist sometimes substitutes a plausible but weaker tool (\texttt{get\_nearby\_agents} where \texttt{get\_agent\_private\_state} answers the need) for one the fixed order reaches reliably. The load-bearing component is therefore gap-triggered budgeting, which we recommend over a fixed probe order (Appendix~\ref{app:tools}) as \sys{}'s deployment configuration: it is the most accurate and cheapest arm we measure. We report the full architecture as primary only for continuity with the rest of the suite, which has no budget-only arm.

\paragraph{Gap calibration: a trigger, not an estimator.}
If the gap score estimated the value of probing, higher-gap queries would benefit more from it. They benefit less. Across the 300 paired cells, per-query benefit correlates negatively with the inferred gap ($r{=}{-}0.33$), falling monotonically as gap rises (bin-wise values in Appendix~\ref{sec:adaptive-budget}). The cause is the subcategory structure already visible above: the inferrer assigns gap almost categorically by query type, and the types it scores highest are the ones a ReAct baseline already probes correctly. What the gap score does reliably is decide whether to cross the probe threshold at all, which the backend ablation shows requires LLM-mediated inference. What it does not do is estimate how much probing will help.

\section{Conclusion}
\looseness=-1
\sys{} inserts an inference step between scene perception and tool use: an \texttt{IntentFrame} whose gap score routes private-state probes before answering, improving implicit-need coverage over ReAct-style probing. Three ablations localise that gain. LLM-mediated gap inference is needed to trigger probing at all; the budget ceiling alone outperforms the full architecture; and gap correlates negatively with the benefit of probing. A small inferred signal deciding \emph{whether} to spend probe budget carries the result, while the frame's finer-grained fields do not yet earn their place. We therefore recommend the leaner controller, and take learning its mapping from logs as the next step.

\section*{Limitations}
\label{sec:analysis}
\label{sec:paradigm}

\paragraph{Regime scope.}
The controller targets situated queries with tool-mediated hidden state. On factual grounding (\S\ref{sec:rq2}) it acts as an access--cost Pareto point rather than an accuracy winner; cross-domain checks on FANToM, LoCoMo, and GAIA show no measurable lift when private state is already in-context or structurally inaccessible (Appendices~\ref{sec:appendix-fantom}--\ref{sec:appendix-gaia}).

\paragraph{Benchmark and calibration.}
The 100-query four-scene benchmark is author-written; on the 25-query pilot subset, inter-annotator agreement over the 5-subcategory partition is substantial ($\kappa{=}0.61$, two independent annotators; the remaining 75 queries carry single-annotator labels, details in Appendix~\ref{sec:appendix-rq-intent-v2}). The few-shot calibration examples are load-bearing for gap estimation: removing them reduces the gain to non-significance, though replacing them with benchmark-disjoint examples preserves it (Appendix~\ref{sec:appendix-prompt-ablation}). The gap score itself is not a calibrated estimator of probe value: bin-wise gap-vs-benefit is inverted ($r{=}{-}0.33$, \S\ref{sec:rq-tom}), and the full architecture underperforms its budget-only ablation ($p{=}0.021$), so claims about the IntentFrame should be read as claims about gap-triggered budget routing, not about the frame's finer-grained fields. Human evaluation uses $N{=}8$ raters (Krippendorff's $\alpha{=}0.43$); directional agreement is strong ($74\%$ consensus) but magnitude estimates carry substantial uncertainty.

\paragraph{Judge circularity.}
\texttt{gpt-4o-mini} serves as both agent backbone and LLM-as-judge for the primary metrics, and this limitation needs two separate answers rather than one. For the implicit-need result, cross-backbone replication (where judge and agent are different models) and a directional human preference are \emph{consistent with} a non-artifact effect but do not establish it; a fully independent judge remains future work. For factual grounding we make no accuracy claim: the strict-precision rescore narrows the margin over ReAct to non-significance ($p{=}0.058$; Appendix~\ref{sec:appendix-strict-rescore}), which does not resolve circularity there. It confirms only that this regime is an access/privacy trade-off, not an accuracy win.

\paragraph{External validity.}
AURATown is one controlled, five-agent, deterministic-private-state setting; the implicit needs are generated by known rules and recovered by probes over the same registry. Nothing in this paper is evidence that the mechanism generalises beyond such settings, and the null cross-domain results (FANToM, LoCoMo, GAIA) mark where it demonstrably stops. Moving to noisier environments, with contradictory private state, tool failures, multiple users, or adversarial privacy pressure, is the necessary next test rather than an incremental extension.

\paragraph{Cross-backbone.}
Three of four tested backbones reproduce the gain; \texttt{gemini-2.5-flash} fails the IntentFrame JSON parser and silently falls back to a heuristic (Appendix~\ref{sec:appendix-cross-backbone}).

\section*{Acknowledgments}
This work is supported by the High-Level Talents Innovation Team Project of the Guangdong--Macao In-Depth Cooperation Zone in Hengqin (Grant No.~2630004018939) and the National Natural Science Foundation of China (NSFC) under Grant No.~62506084.

We thank the three anonymous ARR reviewers and the Area Chair, whose questions during the discussion period led directly to the component ablation and gap-calibration analysis reported in Section~\ref{sec:rq-tom}, results that changed how we describe the mechanism. We also thank the two graduate-student annotators who produced the implicit-intent subcategory labels, and the eight volunteers who completed the human evaluation.

\bibliography{references}

\appendix
\renewcommand{\thesection}{A\arabic{section}}
\renewcommand{\thesubsection}{\thesection.\arabic{subsection}}
\section{Bounded-Probing Algorithm}
\label{app:algorithm}

\begin{algorithm}[H]
\caption{Bounded Proactive Probing (Section~\ref{sec:probing})}
\label{alg:probe}
\begin{algorithmic}[1]
\REQUIRE Agent state $s$, tool registry $\mathcal{T} = \{t_1, \dots, t_K\}$, max steps $B$, LLM planner $\phi$
\ENSURE Probe result $P = (\text{summary}, \text{trace})$
\STATE $\text{context} \gets \textsc{BuildContext}(s)$ \COMMENT{agent, location, time, nearby}
\STATE $\text{trace} \gets \langle\,\rangle$ \COMMENT{empty probe trace}
\FOR{$i = 1$ \TO $B$}
    \STATE $d \gets \phi(\mathcal{T}, \text{context}, \text{trace})$ \COMMENT{LLM decides: call a tool or stop}
    \IF{$d.\text{action} = \texttt{stop}$}
        \STATE \textbf{break}
    \ENDIF
    \STATE $r \gets \mathcal{T}.\textsc{Execute}(d.\text{tool}, d.\text{args})$
    \STATE $\text{trace}.\textsc{Append}((d.\text{tool}, d.\text{args}, r))$
    \STATE $\text{context} \gets \textsc{Update}(\text{context}, r)$
\ENDFOR
\STATE $\text{summary} \gets \textsc{Summarize}(\text{trace})$
\RETURN $(\text{summary}, \text{trace})$
\end{algorithmic}
\end{algorithm}

\section{Adaptive-Budget Evidence (Supplementary)}
\label{app:architecture-figure}

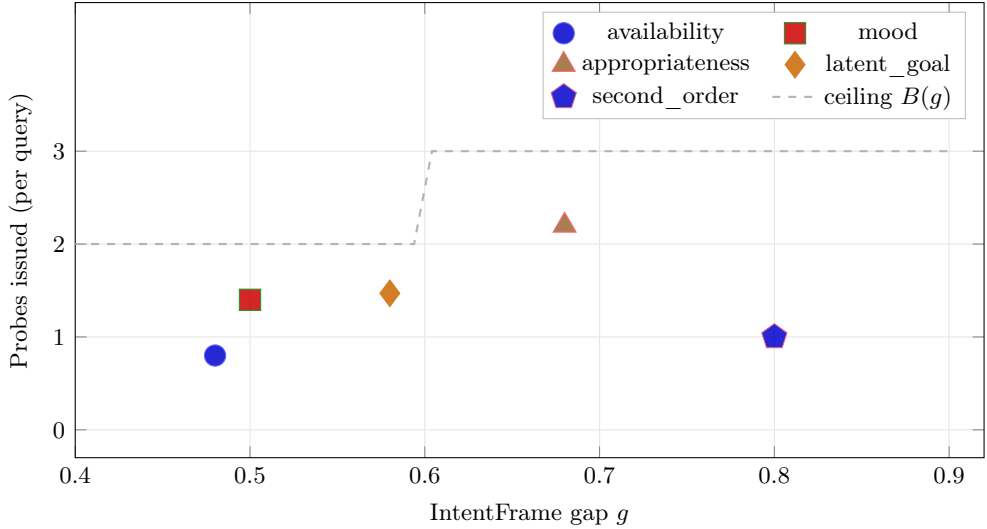
\begin{figure*}[t]
\centering
\begin{tikzpicture}
\begin{axis}[
    width=0.85\textwidth,
    height=6.4cm,
    xlabel={IntentFrame gap $g$},
    ylabel={Probes issued (per query)},
    xmin=0.40, xmax=0.92,
    ymin=-0.3, ymax=3.4,
    xtick={0.4, 0.5, 0.6, 0.7, 0.8, 0.9},
    ytick={0,1,2,3},
    xlabel style={font=\small},
    ylabel style={font=\small},
    xticklabel style={font=\small},
    yticklabel style={font=\small},
    legend style={font=\footnotesize, at={(0.5,-0.22)}, anchor=north, fill=white, draw=gray!50, legend columns=3, /tikz/every even column/.append style={column sep=6pt}},
    grid=major,
    grid style={gray!20},
]
\addplot+[only marks, mark=*, mark size=4pt, color=blue!70, opacity=0.85]
    coordinates {(0.48, 0.80)};   
\addplot+[only marks, mark=square*, mark size=4pt, color=green!50!black, opacity=0.85]
    coordinates {(0.50, 1.40)};   
\addplot+[only marks, mark=triangle*, mark size=5pt, color=red!70, opacity=0.85]
    coordinates {(0.68, 2.20)};   
\addplot+[only marks, mark=diamond*, mark size=5pt, color=orange!80!black, opacity=0.85]
    coordinates {(0.58, 1.47)};   
\addplot+[only marks, mark=pentagon*, mark size=5pt, color=purple!70, opacity=0.85]
    coordinates {(0.80, 1.00)};   
\legend{availability, mood, appropriateness, latent\_goal, second\_order}
\addplot[domain=0.4:0.9, samples=50, dashed, gray!60, thick]
    {(x>=0.8) ? 3 : (x>=0.6 ? 3 : (x>=0.4 ? 2 : (x>=0.2 ? 1 : 0)))};
\addlegendentry{ceiling $B(g)$}
\end{axis}
\end{tikzpicture}
\caption{Per-subcategory mean of \texttt{IntentFrame.gap} vs.\ probes actually issued by the LLM (5 queries $\times$ 3 seeds per marker; \texttt{explore\_max\_steps}=3). The dashed line is the deterministic ceiling map $B(g)$ from Section~\ref{sec:method-intent}, capped at the configured ceiling 3. Probe count varies $2.8\times$ (0.80--2.20) under identical configuration; all four budget values $\{0,1,2,3\}$ appear; Pearson $r(g, \text{probes}){=}0.19$. The \emph{second\_order} subcategory has the highest gap but issues only 1 probe because one targeted \texttt{get\_agent\_belief\_about} call already returns the belief.}
\label{fig:adaptive}
\end{figure*}

\begin{figure}[t]
\centering
\resizebox{\columnwidth}{!}{%
\begin{tikzpicture}
\begin{axis}[
    width=0.92\columnwidth,
    height=4.4cm,
    ybar=2pt,
    bar width=4.0pt,
    enlarge x limits=0.12,
    ymin=0, ymax=1.05,
    ylabel={Implicit Score},
    symbolic x coords={availability, mood, appropriateness, latent\_goal, second\_order},
    xtick=data,
    xticklabel style={font=\scriptsize, rotate=20, anchor=east},
    yticklabel style={font=\footnotesize},
    ylabel style={font=\footnotesize},
    legend style={font=\footnotesize, at={(0.5,1.05)}, anchor=south, legend columns=3, /tikz/every even column/.append style={column sep=4pt}},
    grid=major,
    grid style={gray!20},
]
\addplot+[fill=gray!40, draw=gray!60] coordinates {
    (availability, 0.27)
    (mood, 0.25)
    (appropriateness, 0.35)
    (latent\_goal, 0.00)
    (second\_order, 0.08)
};
\addplot+[fill=blue!30, draw=blue!60] coordinates {
    (availability, 0.28)
    (mood, 0.64)
    (appropriateness, 0.57)
    (latent\_goal, 0.69)
    (second\_order, 1.00)
};
\addplot+[fill=red!50, draw=red!70] coordinates {
    (availability, 0.79)
    (mood, 0.81)
    (appropriateness, 0.84)
    (latent\_goal, 0.81)
    (second\_order, 0.95)
};
\legend{Literal, NoIntent (ReAct-style), \sys{} Intent}
\end{axis}
\end{tikzpicture}}
\caption{Per-subcategory implicit-need surfacing on the implicit-intent benchmark (25 queries, 3 seeds, $N{=}15$ per cell). \sys{} Intent improves most on lexically decoupled private-state queries (e.g.\ \emph{availability}); on \emph{second\_order} queries (``does X think Y?'') the surface already cues belief retrieval and NoIntent ties.}
\label{fig:subcategory}
\end{figure}
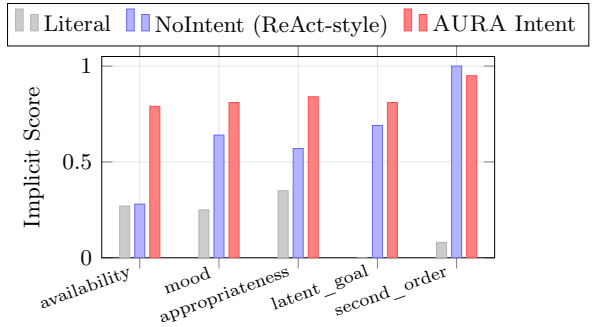

This appendix expands the adaptive-budget claim summarised in Section~\ref{sec:rq-tom}: under a fixed global \texttt{explore\_max\_steps}=3 ceiling, the per-query probe count actually issued by \sys{} varies across subcategories by $2.8\times$, and is driven by the \texttt{IntentFrame}'s gap field rather than the configured ceiling. Figure~\ref{fig:subcategory} shows the per-subcategory implicit-need scores, Figure~\ref{fig:adaptive} the gap-vs-probes scatter, and Table~\ref{tab:adaptive-budget} the per-subcategory means with paired latencies.

\subsection{Adaptive Budget: LLM-Directed Probe Allocation}
\label{sec:adaptive-budget}

A practical concern for any bounded-probing system is whether the
configured probe ceiling simply becomes a fixed per-query cost. We
therefore measure the \emph{effective} number of probes issued after
the \texttt{IntentFrame} sets a ceiling and the Explore loop decides
whether to stop early. This analysis does not try to define
autonomous-agent behavior; it checks the narrower mechanism claim that
intent inference changes probe allocation across queries.

Using the same 25-query~$\times$~3-condition~$\times$~3-seed run as
Section~\ref{sec:rq-tom} (for \sys{}-full only; $N=75$), we measure
the probe-count distribution conditional on query subcategory, while
holding \texttt{explore\_max\_steps=3} constant across all runs. The
configured budget is therefore a ceiling, not a target.

\begin{table}[t]
\centering
\caption{Per-subcategory distributions under \sys{} Full (fixed
\texttt{explore\_max\_steps}=3, $N=15$ per row: 5 queries $\times$ 3 seeds).
Gap is the IntentFrame's literal/implicit divergence; Probes is the
number of tool calls actually issued after early stopping.}
\label{tab:adaptive-budget}
\scriptsize
\setlength{\tabcolsep}{3pt}
\begin{tabular}{@{}lrrrr@{}}
\toprule
\textbf{Subcategory} & \textbf{Gap} & \textbf{Probes} & \textbf{Score} & \textbf{Lat.} \\
\midrule
availability     & $0.48{\pm}0.04$ & $0.80{\pm}0.41$ & $0.79$ & $4.5$ \\
mood             & $0.50{\pm}0.00$ & $1.40{\pm}0.51$ & $0.81$ & $6.0$ \\
appropriateness  & $0.68{\pm}0.06$ & $\mathbf{2.20{\pm}0.68}$ & $0.84$ & $6.1$ \\
latent\_goal     & $0.58{\pm}0.07$ & $1.47{\pm}0.74$ & $0.81$ & $6.0$ \\
second\_order    & $0.80{\pm}0.00$ & $1.00{\pm}0.00$ & $\mathbf{0.95}$ & $5.5$ \\
\bottomrule
\end{tabular}
\end{table}

\paragraph{Observed allocation patterns.}

\emph{(i) Effective probe count is not constant.} Mean probe count per
query ranges from $0.80$ (availability) to $2.20$ (appropriateness),
despite the same global \texttt{explore\_max\_steps}=3 setting.

\emph{(ii) All four budget values are used.} Across the 75 runs, the
agent issued $\{0, 1, 2, 3\}$ tool calls---the full effective range
allowed by the run. This is consistent with the intended design: the
configured budget is a ceiling, not a target.

\emph{(iii) Gap is an input, not the decision itself.} The
\emph{second\_order} subcategory has the highest observed gap ($0.80$)
yet uses only $1.00$ probe on average. The gap-to-budget rule maps
$0.80$ to a ceiling of three probes, but the Explore loop usually
stops after one targeted belief-state call because that call already
returns the needed value. The Pearson correlation between gap and
actual probe count is $r = 0.19$, so the gap is not a proxy for the
final number of tool calls.

Taken together, these traces support the mechanism claim that
\texttt{IntentFrame}-conditioned probing changes the cost and evidence
gathered for different queries under the same run configuration.

\paragraph{Scope of the claim.}
The architecture remains a hybrid system: Perceive, Scene, and Memory
are code-determined, while intent inference, probe selection, and
response generation use LLM calls. The adaptive-budget result should
therefore be read as evidence about one control point in the system,
not as a broad claim about every stage of the pipeline.

\paragraph{Gap-vs-benefit calibration (bin-wise).}
Section~\ref{sec:rq-tom} reports that the gap score is a probe trigger
rather than a calibrated estimate of probe value. Table~\ref{tab:gap-bins}
gives the bin-wise evidence: over the 300 paired query--seed cells of the
100-query benchmark, the mean benefit of intent-directed probing
(\texttt{tom} $-$ \texttt{no\_intent} implicit score) falls monotonically as
the inferred gap rises, and the highest-gap bin is negative. Pearson
$r{=}{-}0.33$, Spearman $-0.32$. Regenerate with
\texttt{scripts/gap\_calibration\_analysis.py}.

\begin{table}[ht]
\centering
\caption{Mean benefit of intent-directed probing by inferred-gap bin
($n{=}300$ query--seed cells; 95\% CI from the cell distribution).}
\label{tab:gap-bins}
\small
\begin{tabular}{lccc}
\toprule
\textbf{Gap bin} & $n$ & \textbf{Mean benefit} & \textbf{95\% CI} \\
\midrule
$[0.4, 0.5)$ & $108$ & $+0.167$ & $[+0.116, +0.217]$ \\
$[0.5, 0.6)$ & $\phantom{0}84$ & $+0.048$ & $[+0.004, +0.092]$ \\
$[0.6, 0.7)$ & $\phantom{0}77$ & $+0.010$ & $[-0.042, +0.063]$ \\
$[0.7, 0.8]$ & $\phantom{0}31$ & $-0.045$ & $[-0.096, +0.005]$ \\
\bottomrule
\end{tabular}
\end{table}

\begin{table}[!htbp]
\centering
\caption{Per-subcategory implicit score ($n{=}15$ per cell, 5 queries $\times$ 3 seeds). Intent inference's marginal gain concentrates on availability queries (surface hides implicit need) and ties NoIntent on second-order queries (surface cues ``X thinks Y'').}
\label{tab:rq-tom-subcat}
\footnotesize
\setlength{\tabcolsep}{4pt}
\resizebox{\columnwidth}{!}{%
\begin{tabular}{@{}lccc@{}}
\toprule
\textbf{Subcategory} & \textbf{Literal} & \textbf{NoIntent} & \textbf{\sys{} Intent} \\
\midrule
availability     & $0.27$ & $0.28$ & $\mathbf{0.79}$ \\
mood             & $0.25$ & $0.64$ & $\mathbf{0.81}$ \\
appropriateness  & $0.35$ & $0.57$ & $\mathbf{0.84}$ \\
latent\_goal     & $0.00$ & $0.69$ & $\mathbf{0.81}$ \\
second\_order    & $0.08$ & $\mathbf{1.00}$ & $0.95$ \\
\bottomrule
\end{tabular}
}
\end{table}

\section{Private-State Evolution Rules}
\label{app:private-state-rules}

Private fields on each agent (\texttt{availability}, \texttt{emotional\_state}, \texttt{unspoken\_goal}, \texttt{beliefs\_about\_others}) update each simulation tick under a deterministic, transparent rule table (source: \texttt{demo/\allowbreak{}town/\allowbreak{}private\_state\_evolution.py}). The intent is to make AURATown's private state a live function of agent action and environment context rather than a static lookup. Rules are evaluated in order; the first match wins.

\begin{table}[t]
\centering
\caption{Private-state evolution rules (first match wins). Action keywords match against \texttt{current\_action} via simple regex; workplace map: Lin Wei$\to$Sunrise Cafe, Chen Mei$\to$General Store, Zhang Hao$\to$home/library, Liu Yang/Wang Jun$\to$Library.}
\label{tab:private-state-rules}
\scriptsize
\setlength{\tabcolsep}{3pt}
\begin{tabular}{@{}p{1.85cm}p{2.55cm}p{2.55cm}@{}}
\toprule
\textbf{Rule} & \textbf{Trigger} & \textbf{Outcome} \\
\midrule
sleep & action matches \texttt{sleep} & DND / resting / no goal \\
workplace\_\allowbreak{}loaded & at workplace, $\geq 3$ peers, busy-action keyword & busy / tired-focused / ``close out the rush'' \\
deep\_focus & action matches \texttt{writ|draft|stud|\allowbreak{}research|meditat} & DND / creatively-flowing / writing-milestone \\
workplace\_\allowbreak{}empty & at workplace, 0 peers, in opening hours & available / lonely / ``hoping a regular drops by'' \\
relaxed & action matches \texttt{walk|read|\allowbreak{}tai chi|eat|sleep} & available; emo recovers from stress, persists from loneliness \\
recent\_stress & last 8 events contain \texttt{argument|\allowbreak{}emergency|failed|\allowbreak{}broke|worried} & available / stressed \\
default & none of the above & available / neutral \\
\bottomrule
\end{tabular}
\end{table}

\paragraph{Beliefs-about-others refresh.}
Each agent's \texttt{beliefs\_about\_others} dictionary entries refresh \emph{only} when this agent is co-located with the target agent in the same simulation tick. Beliefs about non-co-located agents remain at the most recently observed value, so they go stale when peers move. This is the substrate the \emph{second\_order} subcategory of the implicit-intent benchmark (\S\ref{sec:rq-tom}) probes: a query like ``does Lin Wei think Zhang Hao is at home?'' asks for the believer's stale memory, not the target's current ground-truth.

The full rule table is unit-tested at \texttt{tests/\allowbreak{}test\_private\_state\_evolution.py} (14 cases, all rules + co-location/staleness invariants). The evolution itself is pure-function and zero-LLM, so it is reproducible across seeds and is not a confound when ablating other mechanisms.

\section{AURATown Agent Profiles and Map}
\label{app:agents}

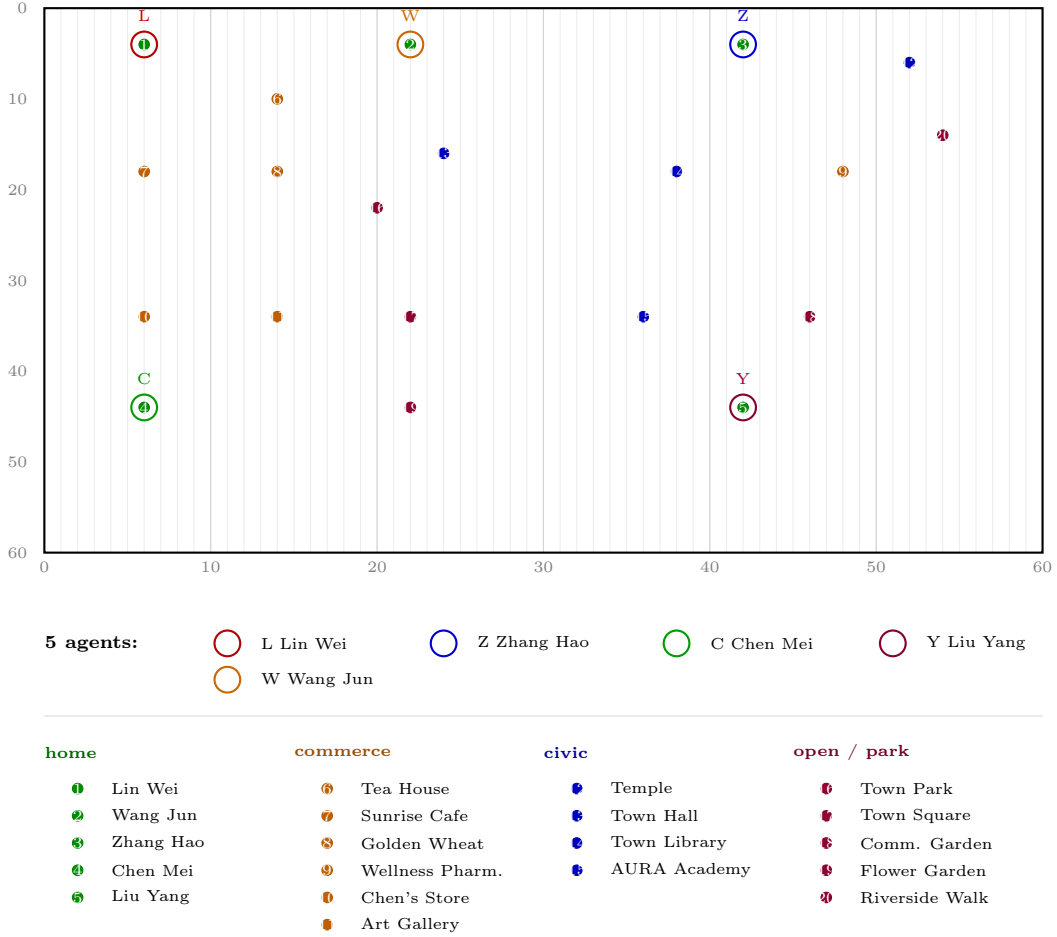
\begin{figure*}[t]
\centering
\begin{tikzpicture}[
    x=0.13cm, y=-0.13cm,
    every node/.style={font=\scriptsize, inner sep=0pt},
    home/.style     ={draw=green!50!black, fill=green!55!black, circle, inner sep=2.3pt},
    commerce/.style ={draw=orange!70!black, fill=orange!75!black, circle, inner sep=2.3pt},
    civic/.style    ={draw=blue!70!black, fill=blue!75!black, circle, inner sep=2.3pt},
    park/.style     ={draw=purple!70!black, fill=purple!75!black, circle, inner sep=2.3pt},
    agent/.style    ={draw=#1, line width=0.8pt, circle, minimum size=3.4mm},
    idlabel/.style  ={font=\tiny, text=white, inner sep=0pt},
    legendpin/.style={font=\tiny, text=white, inner sep=0pt},
]

\draw[gray!15, very thin, step=1]  (0, 0) grid (60, 60);
\draw[gray!35, thin,      step=10] (0, 0) grid (60, 60);
\draw[black, thick] (0, 0) rectangle (60, 60);

\foreach \x in {0, 10, 20, 30, 40, 50, 60}
    \node[font=\tiny, text=gray, anchor=north] at (\x, 61) {\x};
\foreach \y in {0, 10, 20, 30, 40, 50, 60}
    \node[font=\tiny, text=gray, anchor=east] at (-1, \y) {\y};

\node[home] (L1) at ( 6,  4) {}; \node[idlabel] at (L1) {1};
\node[home] (L2) at (22,  4) {}; \node[idlabel] at (L2) {2};
\node[home] (L3) at (42,  4) {}; \node[idlabel] at (L3) {3};
\node[home] (L4) at ( 6, 44) {}; \node[idlabel] at (L4) {4};
\node[home] (L5) at (42, 44) {}; \node[idlabel] at (L5) {5};
\node[commerce] (L6)  at (14, 10) {}; \node[idlabel] at (L6)  {6};
\node[commerce] (L7)  at ( 6, 18) {}; \node[idlabel] at (L7)  {7};
\node[commerce] (L8)  at (14, 18) {}; \node[idlabel] at (L8)  {8};
\node[commerce] (L9)  at (48, 18) {}; \node[idlabel] at (L9)  {9};
\node[commerce] (L10) at ( 6, 34) {}; \node[idlabel, font=\tiny] at (L10) {\tiny 10};
\node[commerce] (L11) at (14, 34) {}; \node[idlabel, font=\tiny] at (L11) {\tiny 11};
\node[civic] (L12) at (52,  6) {}; \node[idlabel, font=\tiny] at (L12) {\tiny 12};
\node[civic] (L13) at (24, 16) {}; \node[idlabel, font=\tiny] at (L13) {\tiny 13};
\node[civic] (L14) at (38, 18) {}; \node[idlabel, font=\tiny] at (L14) {\tiny 14};
\node[civic] (L15) at (36, 34) {}; \node[idlabel, font=\tiny] at (L15) {\tiny 15};
\node[park] (L16) at (20, 22) {}; \node[idlabel, font=\tiny] at (L16) {\tiny 16};
\node[park] (L17) at (22, 34) {}; \node[idlabel, font=\tiny] at (L17) {\tiny 17};
\node[park] (L18) at (46, 34) {}; \node[idlabel, font=\tiny] at (L18) {\tiny 18};
\node[park] (L19) at (22, 44) {}; \node[idlabel, font=\tiny] at (L19) {\tiny 19};
\node[park] (L20) at (54, 14) {}; \node[idlabel, font=\tiny] at (L20) {\tiny 20};

\node[agent=red!70!black]    at ( 6,  4) {}; \node[font=\tiny, anchor=south, text=red!80!black]    at ( 6,  1.5) {L};
\node[agent=orange!80!black] at (22,  4) {}; \node[font=\tiny, anchor=south, text=orange!80!black] at (22,  1.5) {W};
\node[agent=blue!80!black]   at (42,  4) {}; \node[font=\tiny, anchor=south, text=blue!80!black]   at (42,  1.5) {Z};
\node[agent=green!60!black]  at ( 6, 44) {}; \node[font=\tiny, anchor=south, text=green!55!black]  at ( 6, 41.5) {C};
\node[agent=purple!70!black] at (42, 44) {}; \node[font=\tiny, anchor=south, text=purple!80!black] at (42, 41.5) {Y};

\begin{scope}[shift={(66, 2)}]

  \node[font=\scriptsize\bfseries, anchor=west] at (0, 0) {5 agents};
  \node[agent=red!70!black]    at (1.5,  4) {}; \node[font=\tiny, anchor=west] at (4,  4) {L Lin Wei};
  \node[agent=blue!80!black]   at (1.5,  8) {}; \node[font=\tiny, anchor=west] at (4,  8) {Z Zhang Hao};
  \node[agent=green!60!black]  at (1.5, 12) {}; \node[font=\tiny, anchor=west] at (4, 12) {C Chen Mei};
  \node[agent=purple!70!black] at (1.5, 16) {}; \node[font=\tiny, anchor=west] at (4, 16) {Y Liu Yang};
  \node[agent=orange!80!black] at (1.5, 20) {}; \node[font=\tiny, anchor=west] at (4, 20) {W Wang Jun};

  \draw[gray!30, thin] (0, 25) -- (22, 25);

  \node[font=\tiny\bfseries, text=green!40!black, anchor=west] at (0, 29) {home};
  \node[home] at (2, 33) {}; \node[legendpin] at (2, 33) {1}; \node[font=\tiny, anchor=west] at (4.5, 33) {Lin Wei};
  \node[home] at (2, 37) {}; \node[legendpin] at (2, 37) {2}; \node[font=\tiny, anchor=west] at (4.5, 37) {Wang Jun};
  \node[home] at (2, 41) {}; \node[legendpin] at (2, 41) {3}; \node[font=\tiny, anchor=west] at (4.5, 41) {Zhang Hao};
  \node[home] at (2, 45) {}; \node[legendpin] at (2, 45) {4}; \node[font=\tiny, anchor=west] at (4.5, 45) {Chen Mei};
  \node[home] at (2, 49) {}; \node[legendpin] at (2, 49) {5}; \node[font=\tiny, anchor=west] at (4.5, 49) {Liu Yang};

  \node[font=\tiny\bfseries, text=orange!60!black, anchor=west] at (28, 0) {commerce};
  \node[commerce] at (30,  4) {}; \node[legendpin] at (30,  4) {6};        \node[font=\tiny, anchor=west] at (32.5,  4) {Tea House};
  \node[commerce] at (30,  8) {}; \node[legendpin] at (30,  8) {7};        \node[font=\tiny, anchor=west] at (32.5,  8) {Sunrise Cafe};
  \node[commerce] at (30, 12) {}; \node[legendpin] at (30, 12) {8};        \node[font=\tiny, anchor=west] at (32.5, 12) {Golden Wheat};
  \node[commerce] at (30, 16) {}; \node[legendpin] at (30, 16) {9};        \node[font=\tiny, anchor=west] at (32.5, 16) {Pharmacy};
  \node[commerce] at (30, 20) {}; \node[legendpin] at (30, 20) {\tiny 10}; \node[font=\tiny, anchor=west] at (32.5, 20) {Chen's Store};
  \node[commerce] at (30, 24) {}; \node[legendpin] at (30, 24) {\tiny 11}; \node[font=\tiny, anchor=west] at (32.5, 24) {Art Gallery};

  \node[font=\tiny\bfseries, text=blue!60!black, anchor=west] at (28, 29) {civic};
  \node[civic] at (30, 33) {}; \node[legendpin] at (30, 33) {\tiny 12}; \node[font=\tiny, anchor=west] at (32.5, 33) {Temple};
  \node[civic] at (30, 37) {}; \node[legendpin] at (30, 37) {\tiny 13}; \node[font=\tiny, anchor=west] at (32.5, 37) {Town Hall};
  \node[civic] at (30, 41) {}; \node[legendpin] at (30, 41) {\tiny 14}; \node[font=\tiny, anchor=west] at (32.5, 41) {Town Library};
  \node[civic] at (30, 45) {}; \node[legendpin] at (30, 45) {\tiny 15}; \node[font=\tiny, anchor=west] at (32.5, 45) {AURA Academy};

  \node[font=\tiny\bfseries, text=purple!60!black, anchor=west] at (28, 50) {open / park};
  \node[park] at (30, 54) {}; \node[legendpin] at (30, 54) {\tiny 16}; \node[font=\tiny, anchor=west] at (32.5, 54) {Town Park};
  \node[park] at (30, 58) {}; \node[legendpin] at (30, 58) {\tiny 17}; \node[font=\tiny, anchor=west] at (32.5, 58) {Town Square};
  \node[park] at (30, 62) {}; \node[legendpin] at (30, 62) {\tiny 18}; \node[font=\tiny, anchor=west] at (32.5, 62) {Comm. Garden};
  \node[park] at (30, 66) {}; \node[legendpin] at (30, 66) {\tiny 19}; \node[font=\tiny, anchor=west] at (32.5, 66) {Flower Garden};
  \node[park] at (30, 70) {}; \node[legendpin] at (30, 70) {\tiny 20}; \node[font=\tiny, anchor=west] at (32.5, 70) {Riverside Walk};
\end{scope}

\end{tikzpicture}
\caption{\textsc{AURATown} --- 60$\times$60 grid with 20 named locations at their real coordinates (from \texttt{demo/town/assets/town\_map.json}), coloured by type (\textcolor{green!55!black}{home} 5, \textcolor{orange!75!black}{commerce} 6, \textcolor{blue!75!black}{civic} 4, \textcolor{purple!75!black}{open/park} 5). Rings labelled L/Z/C/Y/W mark the five agents at their starting homes per \texttt{demo/town/agents.py}. Public state (location, action, nearby agents) is visible in the scene snapshot; private state (\texttt{availability}, \texttt{emotional\_state}, \texttt{unspoken\_goal}, \texttt{beliefs\_about\_others}) is hidden and only retrievable via the probe tools listed in Table~\ref{tab:tools}.}
\label{fig:auratown}
\end{figure*}

\textsc{AURATown} is a 60$\times$60-grid simulation of five named agents living in twenty named locations (homes, commerce, civic, parks). The five agents (Table~\ref{tab:agents}) and their starting locations (Figure~\ref{fig:auratown}) are fixed across all experiments. Each agent has both public state (location, current action, nearby agents) visible to all queries and private state (\texttt{availability}, \texttt{emotional\_state}, \texttt{unspoken\_goal}, \texttt{beliefs\_about\_others}) accessible only via probe tools.

\begin{table}[t]
\centering
\caption{Agent profiles in \textsc{AURATown}.}
\label{tab:agents}
\footnotesize
\setlength{\tabcolsep}{4pt}
\begin{tabular}{@{}llp{2.0cm}p{2.4cm}@{}}
\toprule
\textbf{Name} & \textbf{Age} & \textbf{Occupation} & \textbf{Personality} \\
\midrule
Lin Wei & 32 & Cafe Owner & Warm, social \\
Zhang Hao & 28 & Writer & Introverted, observant \\
Chen Mei & 45 & Shop Owner & Practical, connector \\
Liu Yang & 20 & Student & Curious, idealistic \\
Wang Jun & 68 & Retired Prof. & Wise, mentor-like \\
\bottomrule
\end{tabular}
\end{table}

\section{Probe Tool Descriptions}
\label{app:tools}

Factual grounding (\S\ref{sec:rq2}) uses an eight-tool environment registry (Table~\ref{tab:tools}); each tool returns a structured fragment of the simulation state. The implicit-intent setup (\S\ref{sec:rq-tom}) uses a separate five-tool scripted registry (\texttt{get\_all\_agents}, \texttt{get\_nearby\_agents}, \texttt{get\_agent\_plan}, \texttt{get\_agent\_private\_state}, \texttt{get\_agent\_belief\_about}) over a fixed scene snapshot rather than event-history. The two registries are disjoint by design --- the factual-grounding registry probes the world state through scene-aware tools, the implicit-intent registry probes individual agents' public and private states. The chat/demo deployment additionally exposes both registries, so a deployed agent has access to both base environment tools and private-state/belief probes.

\begin{table}[t]
\centering
\caption{Environment tools available to the probe planner.}
\label{tab:tools}
\footnotesize
\setlength{\tabcolsep}{3pt}
\begin{tabular}{@{}p{2.65cm}p{1.20cm}p{2.55cm}@{}}
\toprule
\textbf{Tool} & \textbf{Args} & \textbf{Description} \\
\midrule
\texttt{world.time} & -- & Sim time and day \\
\texttt{world.location} & -- & Agent's location details \\
\texttt{world.\allowbreak{}nearby\_agents} & limit & Agents at same location \\
\texttt{world.\allowbreak{}agents\_summary} & limit & All agents' locations \\
\texttt{memory.recent} & limit & Agent's recent memories \\
\texttt{world.\allowbreak{}events\_recent} & limit & Recent global events \\
\texttt{agent.plan} & -- & Agent's current plan \\
\texttt{world.\allowbreak{}location\_info} & location & Named-location details \\
\bottomrule
\end{tabular}
\end{table}

\section{Implementation Details}
\label{app:implementation}

This section records the LLM, memory, simulation, and infrastructure parameters that hold throughout the experiments. All numbers report mean$\pm$std over the three seeds $\{42, 123, 456\}$ unless explicitly noted.

\paragraph{LLM Configuration.}
All LLM calls use \texttt{gpt-4o-mini} via the OpenAI API with:
\begin{itemize}
    \item Action decision: temperature 0.7, default max tokens
    \item Probe planning: temperature 0.2, max tokens 200
    \item Conversation generation: temperature 0.8, max tokens 600
    \item Reflection: temperature 0.5, default max tokens
    \item Importance scoring: temperature 0.1, max tokens 50
\end{itemize}

\paragraph{Memory Configuration.}
Maximum 200 items per agent, retrieval weights $w_r = 0.3, w_p = 0.4, w_v = 0.3$, recency decay $\lambda = 0.01$, reflection threshold $\theta = 10$.

\paragraph{Simulation Parameters.}
30-minute ticks, 6:00--23:00 day cycle, conversation cooldown 2 ticks, probe cooldown 2 ticks, movement speed 3 grid units/tick.

\paragraph{Infrastructure.}
Python backend serving HTTP API on port 7861 with threading for concurrent state/chat/step requests.
React frontend with canvas-based pixel-art rendering, viewport camera system, and real-time state updates.


\section{Concrete Examples, Comparisons, and Diagrams}
\label{app:examples}

This appendix collects the figures and the proactive-frameworks comparison referenced from the body but moved out for space.

\begin{figure}[!htbp]
\centering
\includegraphics[width=0.95\columnwidth]{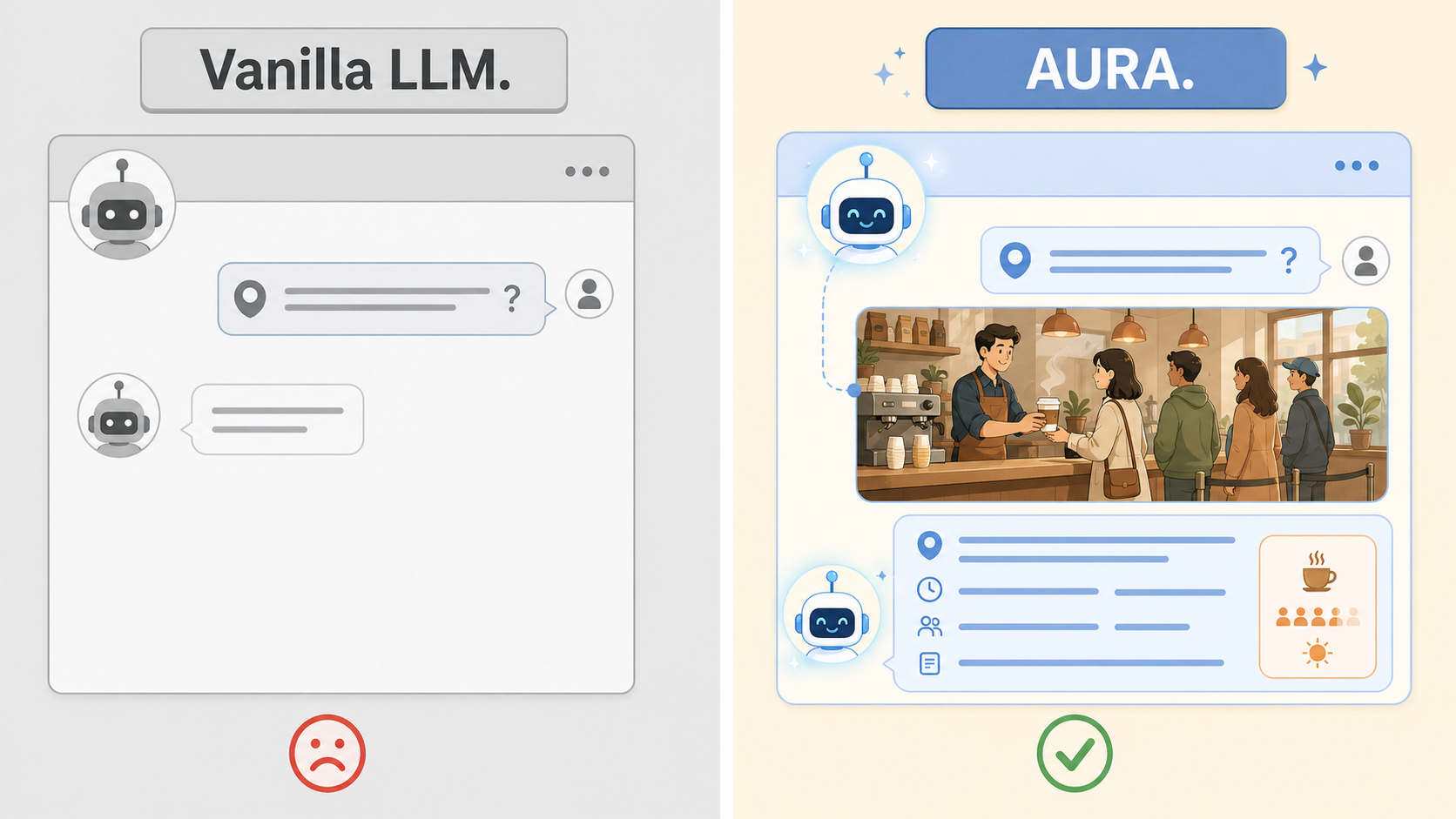}
\caption{Motivating contrast on ``Where is Lin Wei?''. Vanilla (left) vs.\ \sys{} (right) on the same scene state.}
\label{fig:motivating}
\end{figure}

\paragraph{Motivating contrast (Figure~\ref{fig:motivating}).}
\label{app:motivating}
A vanilla LLM and \sys{} on the same query in the same scene state. Vanilla returns the literal location only. \sys{} infers an implicit availability need, probes the cafe scene for nearby agents and the target's private state, and returns the location with a heads-up alert that she is currently busy.

\begin{figure}[!htbp]
\centering
\includegraphics[width=0.95\columnwidth]{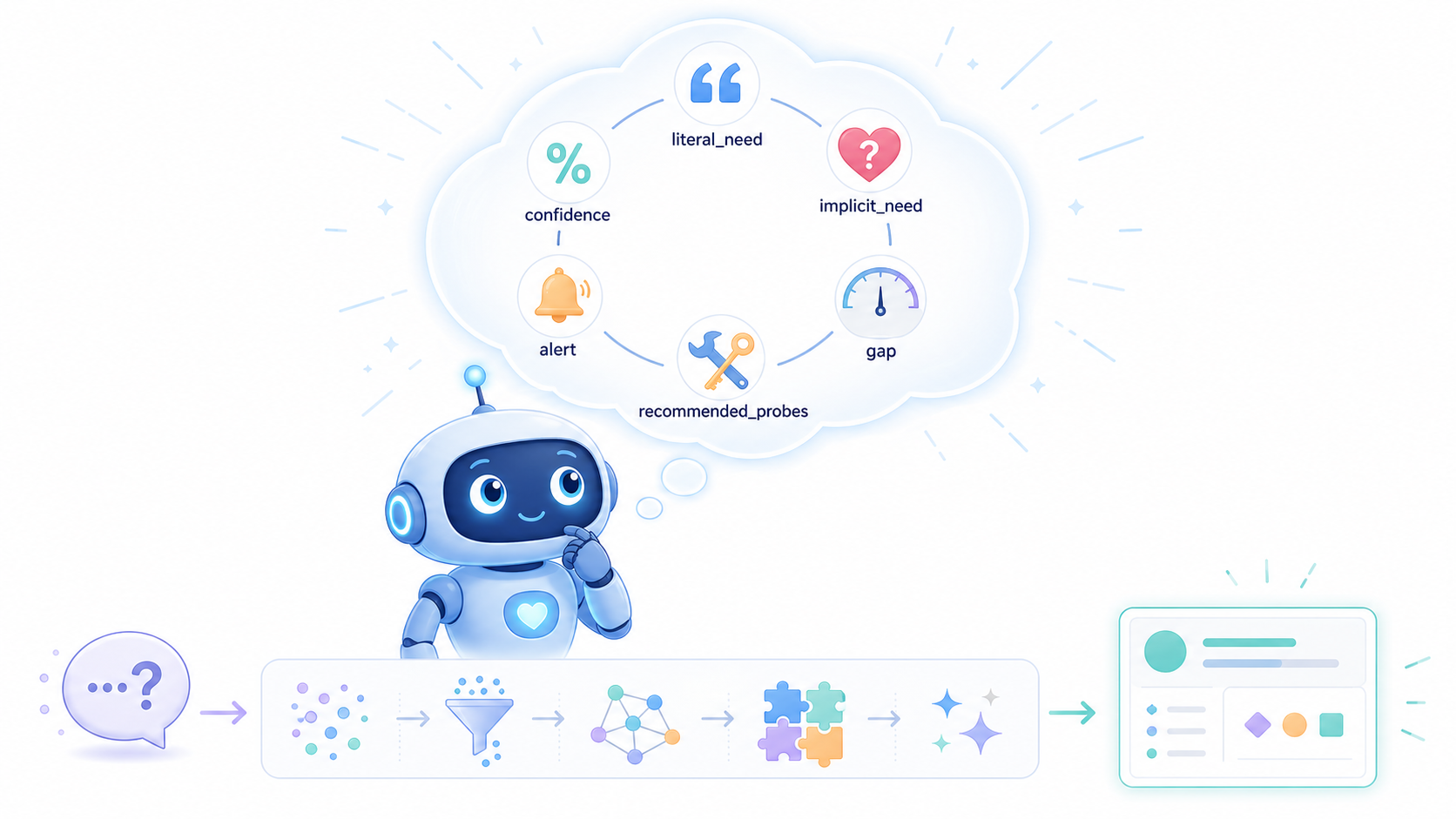}
\caption{The \texttt{IntentFrame} produced for a single user query.}
\label{fig:intentframe}
\end{figure}

\paragraph{IntentFrame example (Figure~\ref{fig:intentframe}).}
\label{app:intentframe-example}
An example \texttt{IntentFrame} produced by the LLM-backed \texttt{IntentInferrer} on a single AURATown query. Six fields are emitted (\texttt{literal\_need}, \texttt{implicit\_need}, \texttt{gap}, \texttt{recommended\_probes}, \texttt{should\_alert}, \texttt{confidence}) and consumed by the Explore and Interact stages (\S\ref{sec:method-intent}).

\paragraph{Long-context prompt-content sketch (Figure~\ref{fig:longcontext}).}
\label{app:longcontext}
Figure~\ref{fig:longcontext} sketches the token-cost difference between packing the full environment state into the prompt and \sys{}'s selective probing on the same \textsc{AURATown} scene. Token counts from the \textsc{AURATown} prompt log; latency and FA from Tables~\ref{tab:hallucination},~\ref{tab:rq-tom-overall}. This is an illustrative sketch, not a head-to-head experiment: the Static-Context baseline (FA $0.450$) is the closest in-suite analogue but does not include private state or memory, so a true full-state long-context baseline is left for future work.

\begin{figure}[t]
\centering
\definecolor{envgreen}{RGB}{40,140,80}
\definecolor{memblue}{RGB}{40,90,170}
\definecolor{warnred}{RGB}{180,60,60}
\begin{tikzpicture}[
  every node/.style={font=\small},
  panel/.style={draw=gray!50, rounded corners=3pt, thick, inner sep=5pt, align=left, text width=\columnwidth-12pt, font=\scriptsize},
  panelhead/.style={font=\small\bfseries},
  pillgreen/.style={draw=envgreen!60!black, fill=envgreen!12, rounded corners=2pt, inner xsep=4pt, inner ysep=2pt, font=\scriptsize},
  pillgray/.style={draw=gray!50, fill=gray!10, rounded corners=2pt, inner xsep=4pt, inner ysep=2pt, font=\scriptsize},
  pillred/.style={draw=warnred!60!black, fill=warnred!10, rounded corners=2pt, inner xsep=4pt, inner ysep=2pt, font=\scriptsize},
  arr/.style={->, thick, gray!60!black, >=stealth},
  query/.style={draw=memblue, fill=memblue!10, rounded corners=2pt, font=\footnotesize, inner sep=4pt},
  ans/.style={draw=envgreen!60!black, fill=envgreen!12, rounded corners=2pt, font=\footnotesize, inner sep=4pt},
]

\node[panel, fill=gray!4] (left) at (0, 0) {
  \textbf{(A) Long-context model}\\[2pt]
  \textit{Pack everything into the prompt; let the LLM attend.}\\[6pt]
  \textbf{Prompt content (per query):}\\
  \texttt{<scene snapshot>} \hfill $\sim$\,180 tok\\
  \texttt{<5 agents $\times$ public state>} \hfill $\sim$\,300 tok\\
  \texttt{<5 agents $\times$ private state>} \hfill $\sim$\,420 tok\\
  \texttt{<recent events / memories>} \hfill $\sim$\,650 tok\\
  \texttt{<system + few-shot>} \hfill $\sim$\,500 tok\\[2pt]
  $\hookrightarrow$ \textbf{$\sim$\,2050 prompt tokens / query}\\
  $\hookrightarrow$ \textbf{O(\#agents)} growth as the world scales \\[6pt]
  \textit{Trade-off:} pays for \emph{all} state every query, even when the
  user's literal need is one location lookup. Long-context attention
  dilution can lower precision on packed prompts.\\[4pt]
  \textbf{Static-Context baseline (factual grounding):} FA $0.450$, $1.8$\,s.
};

\node[panel, fill=envgreen!3, below=0.4cm of left.south west, anchor=north west] (right) {
  \textbf{(B) \sys{} (gap-routed selective probing)}\\[2pt]
  \textit{Infer the gap; fetch only what closes it.}\\[6pt]
  \textbf{Prompt content (per query):}\\
  \texttt{<scene snapshot>} \hfill $\sim$\,180 tok\\
  \texttt{<IntentFrame: literal/implicit/gap>} \hfill $\sim$\,210 tok\\
  \texttt{<probe results $\times$ 1.0$\!\sim\!$2.2>} \hfill $\sim$\,120--280 tok\\
  \texttt{<system + few-shot>} \hfill $\sim$\,500 tok\\[2pt]
  $\hookrightarrow$ \textbf{$\sim$\,1010--1170 prompt tokens / query}\\
  $\hookrightarrow$ \textbf{O(\#probes)} = $0$--$3$, gap-conditioned\\[6pt]
  \textit{Trade-off:} pays one extra LLM call (the IntentInferrer) but skips
  fetching irrelevant private state. Targeted probe returns are more easily
  decoded than a packed prompt.\\[4pt]
  \textbf{\sys{} Full (factual grounding):} FA $0.640$, $3.9$\,s; \\
  \textbf{\sys{} Intent (implicit-intent):} implicit-score $0.803$, $13.8$\,s.
};

\end{tikzpicture}
\caption{\sys{} versus a long-context approach on \textsc{AURATown}. A long-context model pays the full per-agent public+private+memory cost per query (panel A); \sys{}'s \texttt{IntentFrame} routes selective probing so the prompt grows with the inferred gap, not the number of agents (panel B). Token estimates are from the \textsc{AURATown} prompt log; FA / latency from Tables~\ref{tab:hallucination},~\ref{tab:rq-tom-overall}. Static-Context (the long-context lower-bound in our suite) reaches FA $0.450$; \sys{} Full reaches $0.640$.}
\label{fig:longcontext}
\end{figure}

\paragraph{Proactive frameworks comparison (Table~\ref{tab:proactive-comparison}).}
\label{app:proactive-comparison}
Table~\ref{tab:proactive-comparison} positions \sys{} relative to recent proactive and tool-using agent frameworks along three orthogonal axes (decision target, trigger, selection) and reports per-query latency for frameworks we re-implemented under our backbone. ``--'' marks frameworks that address an orthogonal decision (\emph{when to assist} from sensor/behaviour patterns) and were not re-run on our factual benchmark.

\begin{table*}[!tp]
\centering
\caption{Where \sys{} sits among proactive / tool-using agent frameworks. ``Trigger'' = when the framework decides to act; ``Selection'' = how it decides what to fetch; ``Latency'' = mean per-query wall-clock when run on our 50-query factual benchmark with backbone \texttt{gpt-4o-mini}, or ``--'' if not re-implemented here.}
\label{tab:proactive-comparison}
\footnotesize
\setlength{\tabcolsep}{3pt}
\begin{tabular}{@{}lp{2.5cm}p{2.8cm}p{3.2cm}c@{}}
\toprule
\textbf{Framework} & \textbf{Decision target} & \textbf{Trigger} & \textbf{Selection} & \textbf{Latency} \\
\midrule
ContextAgent~\citep{yang2025contextagent}     & when to surface a hint   & sensor stream change          & multi-dim.\ context extract.  & --       \\
ProAgent~\citep{proagent2025sensory}          & predict user need        & AR-glass sensor + profile     & hierarchical context fusion   & --       \\
PROBE~\citep{probe2025proactive}              & search--identify--resolve & idle / unspecified issue      & three-stage pipeline          & --       \\
ProAgentBench~\citep{wang2026proagentbench}   & benchmark (timing+content) & N/A (real-user sessions)    & N/A                           & --       \\
\midrule
ReAct~\citep{yao2023react}                    & during-reasoning act     & per LLM step                  & LLM picks per step            & 6.0\,s   \\
Reflexion~\citep{shinn2023reflexion}          & during + retry           & per step + reflect            & LLM picks per step            & 20.6\,s  \\
Plan-and-Solve~\citep{wang2023plansolve}      & plan-then-execute        & once, from surface query      & surface-query plan            & 5.2\,s   \\
\midrule
\textbf{\sys{} (ours)}                        & \textbf{what to fetch}   & \textbf{post-query, gap-routed} & \textbf{gap $\to$ probe set} & \textbf{3.9\,s} \\
\bottomrule
\end{tabular}
\end{table*}

\section{Experimental Setup (Full)}
\label{sec:appendix-setup}

\paragraph{Baselines.}
We compare \sys{} against six baselines on the factual-grounding benchmark (Table~\ref{tab:hallucination}) and four on the implicit-intent benchmark (Table~\ref{tab:rq-tom-overall}):
\begin{enumerate}[leftmargin=*]
    \item \textbf{Vanilla LLM}: direct \texttt{gpt-4o-mini} generation with no environmental context (factual grounding).
    \item \textbf{Static Context}: LLM receives a packed scene snapshot (location, time, nearby agents) but no proactive information gathering (factual grounding).
    \item \textbf{ReAct Agent}~\citep{yao2023react}: Thought--Action--Observation interleaved \emph{during} reasoning. Same eight base environment tools as \sys{} on factual grounding; same five-tool scripted registry on the implicit-intent benchmark.
    \item \textbf{Reflexion}~\citep{shinn2023reflexion}: ReAct + self-reflection, up to 2 retry rounds (factual grounding).
    \item \textbf{Plan-and-Solve}~\citep{wang2023plansolve}: Phase-1 plan-from-surface-query, Phase-2 execute, Phase-3 synthesise (factual grounding + implicit-intent benchmark).
    \item \textbf{\sys{} (No Probe)}: \sys{} architecture with the probe budget forced to $0$ (factual-grounding ablation).
    \item \textbf{\sys{} (Full)}: complete pipeline with IntentFrame-derived budget and tool selection.
\end{enumerate}
We do not re-run Generative Agents~\citep{park2023generative} or ContextAgent~\citep{yang2025contextagent} as paired baselines: Generative Agents' passive observation string is subsumed by our Static Context; ContextAgent targets a different decision (\emph{when to assist} from sensor streams) and is discussed as related work in Appendix~\ref{app:proactive-comparison} rather than an in-suite comparator.

\paragraph{Models and metrics.}
Backbone \texttt{gpt-4o-mini}, temperature 0.7; \texttt{gpt-4o-mini} LLM-as-judge at temperature 0.1~\citep{zheng2024judging}, combined with a rule-based pre-filter on location and time consistency. Judge and agent share the same model family (gpt-4o-mini), which reduces judge--policy independence; we mitigate by reporting a strict precision rescore (Appendix~\ref{sec:appendix-strict-rescore}) that uses only the judge's CORRECT/CONTRADICTED claim-level classifications and drops the soft completeness component. Metrics: \emph{Grounding Accuracy} (GA, 5 sub-dimensions, inspired by AgentBench~\citep{liu2023agentbench}), \emph{Factual Accuracy} (FA, 50 environment-grounded queries in 5 categories), \emph{SOTOPIA Social Score} (7 dimensions from~\citet{zhou2024sotopia}), \emph{Context Utilization} (CU), \emph{Latency}.

\section{Strict Precision Rescore for Factual Grounding (Robustness Check)}
\label{sec:appendix-strict-rescore}

The factual accuracy (FA) reported in Section~\ref{sec:rq2} uses an LLM-as-judge whose final score combines a hard precision component (correct/(correct+contradicted) over the judge's claim-level classifications, weight $0.7$) with a soft \emph{completeness} component (a 0/0.5/1 rubric for whether the response addressed the question, weight 0.3). Soft components are sensitive to phrasing and judge mood.

We rescore the same per-query records with strict precision only, drop the completeness term, and additionally report the per-query \emph{hallucination rate} (fraction of (query, seed) cells with $\geq 1$ contradicted claim) and \emph{perfect-response rate} (0 contradicted AND $\geq 1$ correct). The rescore script and output JSON are listed in Appendix~\ref{sec:appendix-reproducibility}.

\begin{table}[t]
\centering
\caption{Strict rescore over the same 50 factual-grounding queries $\times$ 3 seeds (per-query, per-seed cells). \emph{StrictP} = correct/(correct + contradicted) excluding cells where the judge produced only unverifiable claims. \emph{Halluc.\,\%} = fraction of cells with $\geq 1$ contradicted claim. \emph{Perfect\,\%} = 0 contradicted AND $\geq 1$ correct. \emph{Lenient FA} = the original 0.7$\cdot$P + 0.3$\cdot$C from Section~\ref{sec:rq2} for comparison.}
\label{tab:rq2-strict}
\small
\resizebox{\columnwidth}{!}{%
\begin{tabular}{lccccc}
\toprule
\textbf{Method} & \textbf{StrictP} & \textbf{Recall} & \textbf{Halluc.\,\%} & \textbf{Perfect\,\%} & \textbf{Lenient FA} \\
\midrule
Vanilla LLM     & $0.036$ & $0.033$ & $93.3$ & $\phantom{0}1.3$ & $0.070$ \\
Static Context  & $0.459$ & $0.396$ & $70.0$ & $26.7$ & $0.450$ \\
ReAct           & $0.578$ & $0.534$ & $51.3$ & $\mathbf{38.0}$ & $0.550$ \\
\sys{} (NoProbe)& $0.581$ & $0.519$ & $75.3$ & $24.0$ & $0.603$ \\
\sys{} (Full)   & $\mathbf{0.658}$ & $\mathbf{0.540}$ & $66.7$ & $28.0$ & $0.640$ \\
\bottomrule
\end{tabular}%
}
\end{table}

Three honest takeaways relative to the lenient table in Section~\ref{sec:rq2}:
\begin{itemize}[leftmargin=*]
\item \textbf{vs.~Vanilla:} the architectural-effect headline holds and gets larger ($+0.609$ strict precision, $p < 10^{-4}$).
\item \textbf{vs.~ReAct (the fair-tools baseline):} the strict precision gap is $+0.090$ ($p{=}0.058$, query-level paired $t$-test, $n{=}49$ complete pairs) --- not significant. \sys{} also has a higher hallucination rate than ReAct ($66.7\%$ vs.\ $51.3\%$, $\Delta{=}+15.3$ pp, $p{=}0.013$). Reading these two together, the architectural pipeline's contribution \emph{above} a tool-using ReAct baseline is small under strict scoring.
\item \textbf{vs.~\sys{} NoProbe:} strict precision rises by $+0.062$ ($p{=}0.032$); the lenient version of this contrast was $p{=}0.299$. The probing mechanism is more clearly supported under strict scoring than under the soft FA, while the overall architectural margin shrinks.
\end{itemize}

The per-category strict precision (Table~\ref{tab:rq2-strict-cat}) localises the gains: \sys{} (Full) wins on temporal ($0.892$) and memory ($0.740$) categories and loses to ReAct on social and spatial. The original ``social win for probing'' from the lenient analysis does not survive strict scoring; we update Section~\ref{sec:rq2}'s scope condition accordingly in the discussion.

\begin{table}[t]
\centering
\caption{Per-category strict precision (mean across query-seed cells, $n{\le}30$ per cell).}
\label{tab:rq2-strict-cat}
\small
\resizebox{\columnwidth}{!}{%
\begin{tabular}{lccccc}
\toprule
\textbf{Category} & \textbf{Vanilla} & \textbf{Static} & \textbf{ReAct} & \textbf{NoProbe} & \textbf{Full} \\
\midrule
memory   & $0.000$ & $0.347$ & $0.547$ & $0.695$ & $\mathbf{0.740}$ \\
planning & $0.034$ & $\mathbf{0.730}$ & $0.453$ & $0.728$ & $0.691$ \\
social   & $0.069$ & $0.275$ & $\mathbf{0.583}$ & $0.291$ & $0.470$ \\
spatial  & $0.034$ & $0.448$ & $\mathbf{0.591}$ & $0.390$ & $0.472$ \\
temporal & $0.036$ & $0.491$ & $0.717$ & $0.798$ & $\mathbf{0.892}$ \\
\bottomrule
\end{tabular}%
}
\end{table}

\section{Privacy-Sensitive Distractor Slice}
\label{sec:appendix-privacy}

The factual-grounding primary FA metric rewards factual completeness but does not penalise unnecessary access. We therefore authored a 30-query factual slice whose gold answers require public facts only. Each query carries a \texttt{forbidden\_tools} list over high-disclosure tools (\texttt{memory.recent}, \texttt{world.events\_recent}, \texttt{agent.plan}) plus query-specific extras; a violation is any fired forbidden tool in the query--seed cell. We run the same three seeds as the factual-grounding benchmark.

\begin{table}[t]
\centering
\caption{Privacy-sensitive factual slice (30 queries $\times$ 3 seeds). \emph{Viol.\,\%} is the fraction of query--seed cells where any forbidden tool fired.}
\label{tab:privacy-distractor}
\small
\resizebox{\columnwidth}{!}{%
\begin{tabular}{lcccc}
\toprule
\textbf{Method} & \textbf{FA} & \textbf{Viol.\,\%} & \textbf{Probes/q} \\
\midrule
Fixed-Probe        & $0.672$ & $100.0$ & $8.00$ \\
ReAct              & $0.618$ & $25.9$  & $1.87$ \\
Plan-and-Solve     & $0.603$ & $78.9$  & $4.16$ \\
\sys{} (GapRouted) & $0.592$ & $\mathbf{0.0}$ & $\mathbf{0.73}$ \\
Static Context     & $0.493$ & $0.0$   & $0.00$ \\
Vanilla LLM        & $0.069$ & $0.0$   & $0.00$ \\
\bottomrule
\end{tabular}%
}
\end{table}

Against GapRouted, Fixed-Probe gains $+0.080$ FA only marginally ($p{=}0.056$) while adding $+100$ pp forbidden-tool violations. Plan-and-Solve and ReAct are tied in FA with GapRouted ($p{=}0.856$ and $p{=}0.655$) but incur $+78.9$ pp and $+25.6$ pp violations. This slice makes the factual-grounding tradeoff explicit: saturated access is often accurate, but it violates stated access constraints by construction; gap routing is not FA-dominant, but it operates on the low-violation, low-probe side of the Pareto frontier.

\section{Cost and Latency Across Conditions}
\label{sec:appendix-cost-latency}

The Pareto framing in Section~\ref{sec:rq2} ranks conditions on access cost (probes) and disclosure, not wall-clock. Table~\ref{tab:cost-latency} reports median and mean per-query latency from the same multi-seed runs so the wall-clock side of the tradeoff is auditable. Medians guard against rare upstream-API timeouts that inflate means (most pronounced for the implicit-intent v1 \emph{tom} row, where one $604$\,s API hiccup on seed $456$ pulls mean to $13.85$\,s while median stays at $5.31$\,s).

\begin{table}[t]
\centering
\caption{Wall-clock per query (seconds; median / mean). Factual columns aggregate the 50q factual benchmark; Privacy is the 30q distractor slice; Implicit-Intent v1 is the 25q implicit-need set under clean prompt. Probe count is the per-condition mean tool-call count (factual / Privacy use the eight-tool registry; implicit-intent uses the five-tool probe registry). \sys{} GapRouted pays an extra IntentInferrer LLM round-trip ($\approx 2$\,s) over a one-shot baseline; the cost does not scale with the probe count.}
\label{tab:cost-latency}
\small
\setlength{\tabcolsep}{4pt}
\resizebox{\columnwidth}{!}{%
\begin{tabular}{lcccccc}
\toprule
 & \multicolumn{2}{c}{\textbf{Factual}} & \multicolumn{2}{c}{\textbf{Privacy slice}} & \multicolumn{2}{c}{\textbf{Implicit-Intent v1}} \\
\cmidrule(lr){2-3}\cmidrule(lr){4-5}\cmidrule(lr){6-7}
\textbf{Condition} & \textbf{Lat} & \textbf{Probes} & \textbf{Lat} & \textbf{Probes} & \textbf{Lat} & \textbf{Probes} \\
\midrule
Vanilla LLM       & $2.14$ / $2.28$ & $0$ & $1.83$ / $2.01$ & $0$ & --- & --- \\
Static Context    & $1.64$ / $1.77$ & $0$ & $1.54$ / $1.59$ & $0$ & --- & --- \\
Literal           & ---             & --- & ---             & --- & $1.23$ / $1.29$ & $0$ \\
NoIntent (ReAct)  & $5.10$ / $6.01$ & $1.05$ & $3.18$ / $4.87$ & $1.87$ & $1.95$ / $2.20$ & $1.05$ \\
Plan-and-Solve    & $4.75$ / $5.19$ & $4.64$ & $4.46$ / $4.51$ & $4.16$ & --- & --- \\
Reflexion         & $16.45$ / $20.64$ & $2.15$ & --- & --- & --- & --- \\
Fixed-Probe       & $2.37$ / $3.23$ & $8.00$ & $1.47$ / $1.65$ & $8.00$ & --- & --- \\
\sys{} GapRouted  & $\mathbf{4.08}$ / $\mathbf{4.34}$ & $\mathbf{1.40}$ & $\mathbf{3.97}$ / $\mathbf{3.99}$ & $\mathbf{0.73}$ & $\mathbf{5.31}$ / $\mathbf{13.85}$\textsuperscript{$\dagger$} & $\mathbf{1.48}$ \\
\bottomrule
\end{tabular}%
}
\end{table}

\noindent\textsuperscript{$\dagger$}\,Median is more representative; mean is inflated by one $604$\,s upstream-API outlier across the $75$ query--seed cells.

Two observations qualify the Pareto picture. First, \sys{} GapRouted is \emph{not} the fastest condition despite firing the fewest probes: the IntentInferrer is a single extra LLM call ($\approx 2$\,s on \texttt{gpt-4o-mini}), so GapRouted's median latency sits above Fixed-Probe ($4.08$ vs.\ $2.37$\,s on factual grounding; $3.97$ vs.\ $1.47$\,s on the privacy slice). The probe-count win does not translate into a wall-clock win at small budgets. Second, Reflexion's median latency ($16.5$\,s) is the regime outlier; its retry loop pays the cost without buying accuracy. We treat latency as a reportable axis rather than a paper claim: the cost-of-selectivity story holds on probe count and disclosure, not on wall-clock.

\section{Prompt Ablation for IntentFrame Calibration}
\label{sec:appendix-prompt-ablation}

We audit whether the \texttt{IntentFrame} gain comes from benchmark-overlapping few-shot examples or from calibrated gap estimation. Table~\ref{tab:prompt-ablation} compares three prompt variants on the same 25 implicit-intent queries and three seeds. The leaked row uses the original benchmark-overlapping examples and is reported only for diagnosis; the clean row is the final system; the no-few-shot row keeps the same rubric but removes all examples.

\begin{table}[!htbp]
\centering
\caption{Prompt ablation for the \texttt{IntentFrame} controller. $p_q$ is the query-level paired test for Intent vs.\ NoIntent.}
\label{tab:prompt-ablation}
\small
\resizebox{\columnwidth}{!}{%
\begin{tabular}{lcccc}
\toprule
\textbf{Variant} & \textbf{Intent} & \textbf{NoIntent} & \textbf{$\Delta$} & \textbf{$p_q$} \\
\midrule
Leaked few-shot & $0.840$ & $0.637$ & $+0.203$ & $7.2{\times}10^{-4}$ \\
Clean few-shot  & $\mathbf{0.803}$ & $0.640$ & $\mathbf{+0.163}$ & $\mathbf{0.006}$ \\
No few-shot     & $0.677$ & $0.643$ & $+0.035$ & $0.44$ \\
\bottomrule
\end{tabular}%
}
\end{table}

Clean few-shot is only $0.037$ below the leaked prompt, so the main effect is not driven by memorising names or locations. In contrast, removing examples reduces Intent to $0.677$ and makes the Intent-vs.-NoIntent gain non-significant. Mechanistically, the no-few-shot prompt underestimates the gap: mean inferred gap drops from $0.589$ to $0.476$, and high-gap cells ($g{\ge}0.60$) drop from $43/75$ to $22/75$. The few-shot examples are therefore load-bearing calibration for gap-to-budget routing.

\section{Cross-Backbone Intent-vs.-NoIntent (Robustness)}
\label{sec:appendix-cross-backbone}

To check that the Intent-vs.-NoIntent gain reported on \texttt{gpt-4o-mini} is not specific to one backbone, we re-ran the 25 implicit-intent queries on three additional production-grade LLMs (each via its vendor's official API; judge fixed at \texttt{gpt-4o-mini} to keep the scoring rubric constant). Table~\ref{tab:rq-tom-backbone} summarises the result. The \texttt{gpt-4o-mini} row is the clean calibrated prompt used in the main paper; the other backbones are single-seed robustness probes. Three of four backbones reproduce the gain; \texttt{gemini-2.5-flash} regresses, but the regression is explained by JSON-schema parse failure on $23/25$ \texttt{IntentFrame} calls and silent fallback to the deterministic heuristic --- a format-compliance boundary.

\begin{table}[!htbp]
\centering
\caption{Cross-backbone Intent-vs.-NoIntent contrast on implicit score (25 implicit-intent queries, query-level paired comparison with cluster bootstrap on \texttt{query\_id}). The \texttt{gpt-4o-mini} row is the clean calibrated prompt and averages over 3 seeds; the other rows are single-seed robustness probes. Three of four backbones reproduce the gain; \texttt{gemini-2.5-flash} fails the JSON parser on $23/25$ \texttt{IntentFrame} calls (run log \texttt{\_run\_gemini25flash.log}) so its row reflects heuristic-fallback under a Gemini tag.}
\label{tab:rq-tom-backbone}
\small
\resizebox{\columnwidth}{!}{%
\begin{tabular}{llccc}
\toprule
\textbf{Backbone} & \textbf{Vendor} & \textbf{NoIntent} & \textbf{\sys{} Intent} & \textbf{$\Delta$ (95\% CI)} \\
\midrule
\texttt{gpt-4o-mini}        & OpenAI    & $0.640$ & $0.803$ & $+0.16$ $[+0.06, +0.27]$ \\
\texttt{claude-haiku-4-5}   & Anthropic & $0.680$ & $\mathbf{0.920}$ & $+0.24$ $[+0.12, +0.36]$ \\
\texttt{qwen-plus}          & Alibaba   & $0.736$ & $\mathbf{0.984}$ & $+0.25$ $[+0.14, +0.37]$ \\
\texttt{gemini-2.5-flash}   & Google    & $0.600$ & $0.400$ & $\mathbf{-0.20}$ $[-0.37, -0.03]$ \\
\bottomrule
\end{tabular}%
}
\end{table}

\section{Expanded Implicit-Intent v2 Details}
\label{sec:appendix-rq-intent-v2}

To test whether the 25-query scene is over-specialised, we authored an expanded implicit-intent v2 set with 4 scene snapshots $\times$ 5 subcategories $\times$ 5 queries ($100$ queries; $300$ scored cells per condition). Scene A preserves the original 25 queries; scenes B--D vary location, time of day, agent rosters, private states, and stale belief-vs-truth mismatches. Each v2 query records \texttt{gold\_required\_tools} and \texttt{forbidden\_tools}; all second-order queries forbid direct \texttt{get\_agent\_private\_state} access because the correct evidence is the believer's recorded belief, not the target's ground truth.
\paragraph{Subcategory examples.}
Representative surface queries by subcategory: \emph{availability} ``where is Lin Wei?'' (implicit: ``is she free?''); \emph{mood} ``how is Chen Mei today?'' (implicit: ``is she in a receptive emotional state?''); \emph{appropriateness} ``is now a good time to invite Lin Wei for coffee?'' (requires integrating schedule and private state); \emph{latent\_goal} ``what is Lin Wei up to?''; \emph{second\_order} ``does Lin Wei think Zhang Hao is free?'' (the correct answer must report the believer's recorded belief, not the target's ground truth

\paragraph{Inter-annotator agreement.}
Two independent annotators (computer-science graduate students, distinct from the authors and naive to the AURA architecture) re-labelled the 25-query pilot set under the 5-subcategory definitions given in the task instructions. They reached Cohen's $\kappa{=}0.61$ (substantial under Landis-Koch; raw agreement $68\%$, $17/25$). All 8 disagreements concentrate on two boundaries: (i) \emph{mood} vs.\ \emph{appropriateness}/\emph{availability} (4/8; e.g., ``Does Zhang Hao look busy?'' admits both an availability reading and a mood reading), and (ii) \emph{appropriateness} vs.\ \emph{literal}/\emph{availability} (3/8; e.g., ``Can I ask Wang Jun for a favor right now?'' splits between a literal-permission reading and a context-aware appropriateness reading). Per-class agreement is highest on \emph{second\_order} (24/25 between annotators) and \emph{latent\_goal} (24/25); collapsing \emph{mood} into a single \emph{context-aware availability} super-category yields $\kappa{=}0.68$. Raw label dumps are released alongside the queries.

\begin{table}[!htbp]
\centering
\caption{Expanded implicit-intent v2 check (100 queries $\times$ 3 seeds, \texttt{gpt-4o-mini}). Scores are implicit-need coverage. The final two rows report \sys{} Intent minus NoIntent; $p$ values are paired tests over query-seed cells.}
\label{tab:rq-intent-v2}
\footnotesize
\setlength{\tabcolsep}{3pt}
\resizebox{\columnwidth}{!}{%
\begin{tabular}{lccccc}
\toprule
 & \textbf{Overall} & \textbf{A: cafe} & \textbf{B: library} & \textbf{C: garden} & \textbf{D: night} \\
\midrule
Literal          & $0.223$ & $0.200$ & $0.264$ & $0.237$ & $0.192$ \\
NoIntent         & $0.733$ & $0.709$ & $0.688$ & $0.731$ & $0.803$ \\
\sys{} Intent    & $0.804$ & $0.800$ & $0.779$ & $0.811$ & $0.827$ \\
\midrule
Intent--NoIntent $\Delta$ & $+0.071$ & $+0.091$ & $+0.091$ & $+0.080$ & $+0.024$ \\
$p$                      & $1.0{\times}10^{-6}$ & $0.004$ & $1.4{\times}10^{-4}$ & $0.015$ & $0.349$ \\
\bottomrule
\end{tabular}%
}
\end{table}

\paragraph{Scene-level read.} Three of four scenes are clearly positive (A/B/C), and the post-event night scene D ties: agents are spatially distributed in D and the public state already telegraphs availability, so the gap mechanism has nothing to add.

\paragraph{Subcategory breakdown (Intent vs.\ NoIntent, cell-level $n{=}60$ per cat).} \emph{availability} $+0.29$ ($p{=}2.7\times10^{-11}$), \emph{appropriateness} $+0.11$ ($p{=}8.2\times10^{-4}$), \emph{mood} $+0.07$ ($p{=}1.7\times10^{-3}$), \emph{second\_order} $-0.02$ ($p{=}0.32$, NS) after a synthesis-prompt fix that drops the public-state dump on belief-state queries and adds a strict ``report belief, not actual state'' instruction, and \emph{latent\_goal} $-0.09$ ($p{=}6.2\times10^{-4}$, residual deficit acknowledged in Limitations).

\paragraph{Cross-backbone v2.} Re-running the 100-query set on \texttt{claude-haiku-4.5} (seed 42 only, via OpenRouter) reproduces the gain at \emph{larger} magnitude: Intent $0.876$ vs.\ NoIntent $0.790$, paired $\Delta{=}{+}0.086$, $p{=}3.7\times10^{-3}$. The scene-D tie pattern replicates ($\Delta{=}{+}0.008$, $p{=}0.88$, NS), confirming the night-scene null is a property of the regime rather than a single-backbone artifact.

\paragraph{A worked failure case (\emph{latent\_goal}).}
The deficit has a consistent shape, visible in a single query. Asked ``is anyone setting up something for tomorrow that I should know about?'', with one agent privately intending to invite another on a hike the next day, \sys{} inferred a high gap, which bought it a full probe ceiling, and then spent that ceiling on a single agent, missing the intent entirely. The NoIntent baseline enumerated the roster first and found it. Naming no subject is what makes such a query hard, and it is also what the shortlist handles worst: a narrow, confident probe plan is a liability when the question ranges over everyone. Across the queries that name no subject, \sys{} probes fewer distinct agents than either the baseline or its own budget-only ablation, so the deficit is one of allocation rather than of budget, and raising the ceiling does not address it.

\section{Factual-Grounding Per-Category Paired Contrast}
\label{sec:appendix-rq2-cat}

The aggregate \sys{} Full vs.\ \sys{} No-Probe contrast on factual grounding is a near-null ($+0.038$ FA, $p{=}0.060$ paired query-level), but this aggregate hides a heterogeneous per-category structure. Splitting the 50 queries into 5 categories of 10 (spatial / social / temporal / memory / planning), Table~\ref{tab:rq2-category} shows that probing produces a significant FA improvement only on the \emph{social} category. The other four categories are saturated by the Perceive/Scene channel, so probing has nothing to add. This is the empirical scope condition for proactive probing referenced in the main text.

\begin{table}[!htbp]
\centering
\caption{Per-category paired $t$-test of proactive probing ($\sys$ Full vs.\ $\sys$ No-Probe), $N{=}10$ queries per category per seed, 3 seeds. Probing produces a significant effect only on the \emph{social} category.}
\label{tab:rq2-category}
\small
\resizebox{\columnwidth}{!}{%
\begin{tabular}{lcccc}
\toprule
\textbf{Category} & \textbf{Full FA} & \textbf{NoProbe FA} & \textbf{$\Delta$} & \textbf{$p$} \\
\midrule
spatial   & $0.457 \pm 0.027$ & $0.453 \pm 0.130$ & $+0.004$ & $0.96$ \\
\textbf{social} & $\mathbf{0.459 \pm 0.056}$ & $\mathbf{0.322 \pm 0.070}$ & $\mathbf{+0.137}$ & $\mathbf{0.010^{\star}}$ \\
temporal  & $0.846 \pm 0.060$ & $0.813 \pm 0.016$ & $+0.033$ & $0.51$ \\
memory    & $0.707 \pm 0.092$ & $0.684 \pm 0.031$ & $+0.023$ & $0.72$ \\
planning  & $0.734 \pm 0.031$ & $0.741 \pm 0.072$ & $-0.007$ & $0.90$ \\
\bottomrule
\end{tabular}%
}
\end{table}

\section{Routine-Action Grounding (Null Result, Full Table)}
\label{sec:appendix-rq1}

The routine-grounding check measures whether richer environmental access changes Grounding Accuracy on routine daily simulation. The protocol is 100 simulation steps $\times$ 5 conditions $\times$ 3 seeds $= 1{,}500$ judgments per condition; the metric averages four GA sub-dimensions (location consistency, time appropriateness, social awareness, plan adherence). Memory utilisation saturates at $\approx 1.0$ for every method and is omitted from Table~\ref{tab:grounding}. The paired $t$-test column reports two-sided $p$ vs.\ Vanilla. All five methods fall within $0.024$ absolute GA spread; every contrast is non-significant. We report this null openly: when most actions are trivially grounded, GA cannot distinguish architectures.

\begin{table}[!htbp]
\centering
\caption{Grounding Accuracy (GA) on 100-step routine daily simulation, 3 seeds, mean $\pm$ std. Dimension scores: location consistency, time appropriateness, social awareness, plan adherence.}
\label{tab:grounding}
\small
\resizebox{\columnwidth}{!}{%
\begin{tabular}{lccccccc}
\toprule
\textbf{Method} & \textbf{GA} & \textbf{Loc.} & \textbf{Time} & \textbf{Social} & \textbf{Plan} & \textbf{Lat.\,(s)} & \textbf{$p$ vs.\ Van.} \\
\midrule
Vanilla LLM           & $0.659 \pm 0.030$ & $0.425$ & $0.371$ & $0.877$ & $0.621$ & $\phantom{0}9.8$  & ---   \\
Static Context        & $0.659 \pm 0.007$ & $0.401$ & $0.384$ & $0.835$ & $0.676$ & $12.9$            & $0.99$ \\
ReAct Agent           & $0.652 \pm 0.015$ & $0.537$ & $0.288$ & $0.911$ & $0.527$ & $29.3$            & $0.73$ \\
\sys{} (No Probe)     & $0.676 \pm 0.008$ & $0.459$ & $0.405$ & $0.837$ & $0.677$ & $16.4$            & $0.51$ \\
\sys{} (Full)         & $0.665 \pm 0.005$ & $0.421$ & $0.375$ & $0.858$ & $0.671$ & $25.2$            & $0.76$ \\
\bottomrule
\end{tabular}%
}
\end{table}

All pairwise paired $t$-tests vs.\ Vanilla give $p > 0.5$; a non-parametric Wilcoxon signed-rank test gives the same conclusion. The GA spread across five different architectures is $0.024$, well inside per-condition seed variance. This is evidence the metric is saturated on this workload, not that probing fails; factual grounding (Section~\ref{sec:rq2}) reaches a $9.1\times$ gap on queries that stress grounding.

\section{Additional Diagnostic Checks}
\label{sec:appendix-minor-rqs}

\subsection{Component Ablation (Saturated-Workload Null)}

We ablate each pipeline component on a routine-action workload of $100$ simulation steps and $50$ chat queries per configuration, three seeds, source data \texttt{\allowbreak{}evaluation/\allowbreak{}\-results/\allowbreak{}rq3\_ablation\_study\_multiseed.json}. Table~\ref{tab:ablation} gives the multi-seed mean GA / FA / latency deltas relative to \sys{} Full. (An earlier draft of this table reported single-seed numbers measured at 20 steps and 20 queries; we replace it here with the full multi-seed configuration to match the actual run.)

\begin{table}[t]
\centering
\caption{Component ablation, multi-seed mean ($\Delta$\ vs.\ \sys{} Full); GA/FA absolute, latency in seconds. Re-computed from \texttt{rq3\_ablation\_study\_multiseed.json} (3 seeds, 100 sim steps, 50 chat queries per config).}
\label{tab:ablation}
\small
\resizebox{\columnwidth}{!}{%
\begin{tabular}{lcccc}
\toprule
\textbf{Configuration} & $\Delta$\ GA & $\Delta$\ FA & $\Delta$\ Lat.\,(s) \\
\midrule
\sys{} Full             & $\phantom{+}0$       & $\phantom{+}0$        & $\phantom{+}0$ \\
w/o Probing             & $+0.002$ & $+0.009$  & $-1.33$ \\
w/o Memory              & $-0.017$ & $+0.026$  & $+0.15$ \\
w/o Reflection          & $-0.003$ & $+0.005$  & $-0.21$ \\
w/o Memory \& Reflect.  & $-0.015$ & $-0.028$  & $+0.30$ \\
Vanilla (all off)       & $-0.020$ & $+0.022$  & $-1.36$ \\
\bottomrule
\end{tabular}%
}
\end{table}

\paragraph{Interpretation.} Component-level deltas on the routine-action workload are uniformly small ($|\Delta\text{GA}| \le 0.020$, $|\Delta\text{FA}| \le 0.028$). Removing the probing component does \emph{not} hurt GA or FA on this workload; it saves $1.3$~s of latency. The same null pattern that drives the routine-grounding saturation result (Section~\ref{sec:appendix-rq1}) drives the component-ablation null: routine actions like ``sleeping at home at 6\,AM'' pass any reasonable grounding check, and the LLM-as-judge factual scoring with completeness has a $\pm 0.03$ noise floor that swamps the per-component contribution. Memory has the largest negative GA effect ($-0.017$) but with FA actually slightly higher when memory is removed --- consistent with the same noise floor, not a clean component effect. The picture changes on the factual-grounding chat workload (Section~\ref{sec:rq2}), where the same architecture moves FA from $0.07$ (Vanilla) to $0.64$ (Full); the bottleneck the ablation in this table fails to expose is the same one the routine-grounding check fails to expose: action-grounding on routine days does not stress the structured-environment-access channel.

\subsection{Emergent Social Behaviours (SOTOPIA)}

200-step multi-agent simulation, 36 conversations evaluated on SOTOPIA's 7 dimensions~\citep{zhou2024sotopia}. Overall quality 7.87/10, strongest dimensions \emph{goal} (9.5) and \emph{believability} (9.0); 44 emergent behaviours across 4 categories: collaboration (32), routine adaptation (7), conflict resolution (4), group formation (1).

\begin{table}[t]
\centering
\caption{SOTOPIA dimension averages (\sys{} full run, 200 sim steps).}
\label{tab:sotopia}
\small
\begin{tabular}{@{}lr@{}}
\toprule
\textbf{Dimension} & \textbf{Avg.\ (range)} \\
\midrule
believability   & $9.0\phantom{0}$ (0--10) \\
goal            & $9.5\phantom{0}$ (0--10) \\
knowledge       & $8.0\phantom{0}$ (0--10) \\
relationship    & $2.07$ ($-5$--$5$) \\
financial       & $0.63$ ($-5$--$5$) \\
secret          & $-0.67$ ($-10$--$0$) \\
social rules    & $-2.13$ ($-10$--$0$) \\
\midrule
overall quality & $7.87$ \\
\bottomrule
\end{tabular}
\end{table}

\begin{figure}[!htbp]
\centering
\includegraphics[width=0.65\columnwidth]{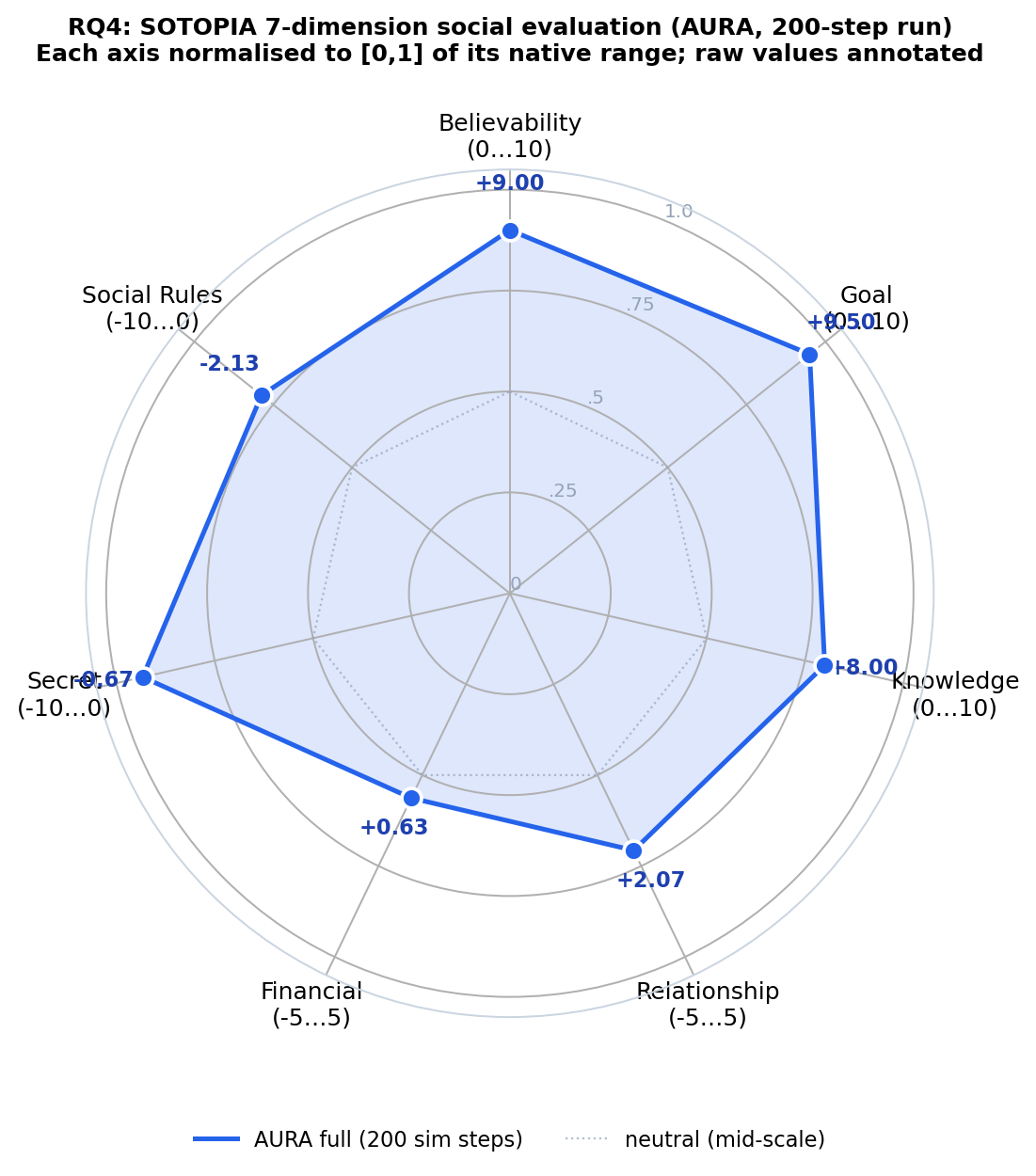}
\caption{SOTOPIA 7-dimension social evaluation (\sys{} full, 200-step run). Each axis is normalised to $[0,1]$ of its native range (shown next to each label); raw values annotated. Strongest dimensions: \emph{believability} ($9.0/10$), \emph{goal} ($9.5/10$); weakest: \emph{social rules} ($-2.13$) and \emph{secret} ($-0.67$) (both negative-only scales). The shape is \emph{lopsided-positive}: \sys{} agents excel on individual-level dimensions and underperform on multi-agent normative dimensions, consistent with the per-agent-mechanism scope of the design.}
\label{fig:rq4-sotopia-radar}
\end{figure}

The collective-behaviour analysis of these numbers is in Appendix~\ref{sec:appendix-extra} (Section~\ref{sec:collective-tom}).

\subsection{Human Evaluation}
\label{sec:rq5}

We collected pairwise A/B annotations from 8 independent raters on
the 50 chat scenarios, four dimensions each (\emph{response
helpfulness}, \emph{environmental awareness}, \emph{agent
believability}, \emph{factual accuracy}) on a 5-point Likert scale,
side-randomised and blinded to system identity. This gives $N{=}400$
paired query-rater observations per dimension. Both responses on
each item were generated with the fixed system prompt at the same
\textsc{AURATown} simulation tick (warmup 10 steps, seed 42); raters
saw the query, the asking agent's name, the category, and the two
anonymous responses but not the underlying simulation scene state.

\begin{table}[t]
\centering
\caption{Human eval primary analysis: rater-aggregated paired Wilcoxon. Each rater's $\Delta$ is averaged across the 50 scenarios first; the test is then over $N{=}8$ paired rater-means per dimension. This is conservative relative to a 400-cell paired test, which would treat repeated-measures within a rater as independent. The 95\% CI column is a 5000-resample cluster bootstrap on \texttt{rater\_id} over the raw $\Delta$ values.}
\label{tab:rq5}
\small
\resizebox{\columnwidth}{!}{%
\begin{tabular}{lccccc}
\toprule
\textbf{Dimension} & \textbf{\sys{}} & \textbf{Vanilla} & \textbf{$\Delta$ (mean of rater means)} & \textbf{Wilcoxon $p$ ($n{=}8$)} & \textbf{Cluster CI} \\
\midrule
Response Helpfulness        & $3.83$ & $2.25$ & $+1.58$ & $0.017$ & $[+0.94, +2.20]$ \\
Environmental Awareness     & $3.82$ & $1.96$ & $\mathbf{+1.86}$ & $0.017$ & $\mathbf{[+1.08, +2.62]}$ \\
Agent Believability         & $3.75$ & $2.16$ & $+1.59$ & $0.017$ & $[+0.88, +2.32]$ \\
Factual Accuracy            & $3.71$ & $2.32$ & $+1.39$ & $0.017$ & $[+0.76, +1.90]$ \\
\bottomrule
\end{tabular}%
}
\end{table}

\begin{figure}[!htbp]
\centering
\includegraphics[width=\columnwidth]{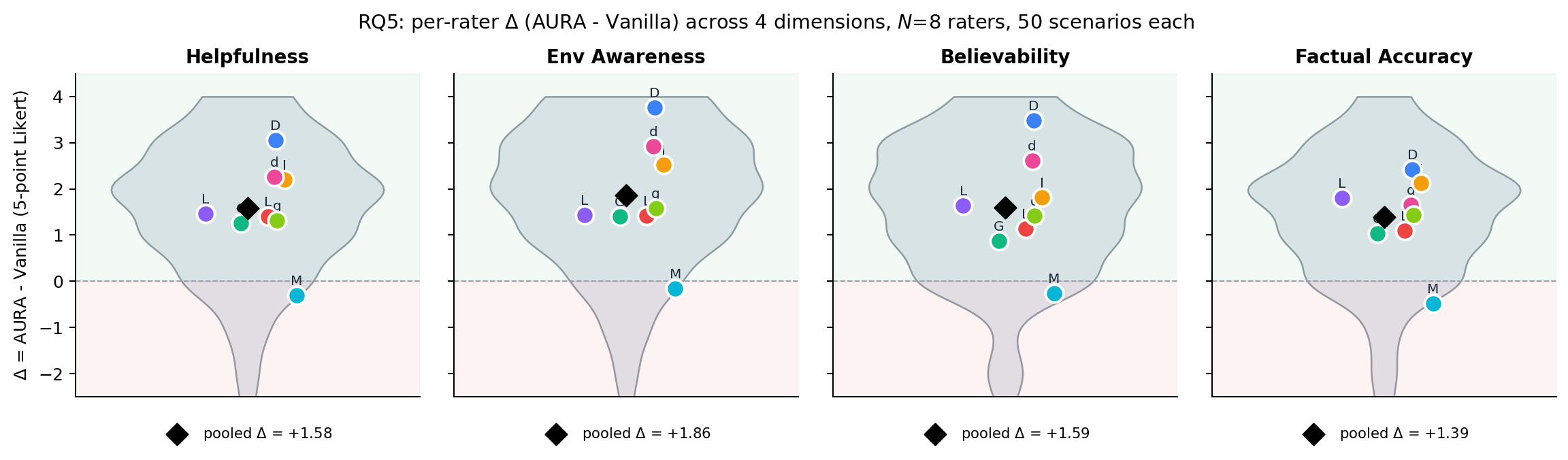}
\caption{Per-rater $\Delta$ (\sys{} $-$ Vanilla) across the 4 dimensions ($N{=}8$ raters, 50 scenarios each). Each coloured dot is one rater's mean $\Delta$ across the 50 scenarios; the grey violin is the pooled $\Delta$ distribution; the black diamond is the pooled mean. Letters identify raters ($M$ is the one whose $\Delta$ falls below zero on factual accuracy and is the basis for the $N{=}7$ sensitivity analysis). 7/8 raters' means lie above zero on every dimension.}
\label{fig:rq5-violin}
\end{figure}

\paragraph{Headline.}
\sys{} receives higher ratings than Vanilla LLM on all four dimensions under the conservative rater-aggregated test ($N{=}8$ paired rater-means per dimension; Wilcoxon $p{=}0.017$ each, sign 7-1-0 for every dimension). All four cluster-bootstrap CIs on the per-rater $\Delta$ exclude zero. The largest gap is on \emph{environmental awareness} ($\Delta{=}+1.86$, CI $[+1.08, +2.62]$), consistent with the environment-mediated design. The cell-level paired test on the 400 (rater $\times$ scenario) cells gives $p<10^{-4}$ but treats within-rater repeated measures as independent, so we report the rater-aggregated $p{=}0.017$ as the primary statistic. Fig.~\ref{fig:rq5-violin} visualises the per-rater means against the pooled distribution.

\paragraph{Per-rater profile (transparency).}
Per-rater mean $\Delta$ ranged from $-0.30$ to $+3.18$ across the 8
raters; per-rater AURA-better rates were
$\{24.0, 69.0, 81.5, 82.5, 85.0, 85.0, 98.5, 100.0\}\%$. Seven of
eight raters preferred \sys{} on aggregate; one rater (24\% AURA-wins,
$\Delta{=}-0.30$) reported in post-task debrief that they had
penalised plausible-but-unverifiable specifics in \sys{}'s responses
as factual errors. We retain that rater in the primary analysis to
avoid post-hoc exclusion bias and discuss the implication under
``Methodological limitations'' below.

\paragraph{Direction-agreement at the cell level.}
For each of the 200 (scenario, dimension) cells, we count raters
preferring \sys{}, Vanilla, or scoring tied. Defining ``consensus''
as $\ge 6$ of 8 raters agreeing on direction, we find:
$\mathbf{0/200}$ cells have a Vanilla consensus,
$\mathbf{148/200}$ ($74\%$) have an \sys{} consensus, and $52$ are
split. Per-dimension \sys{}-consensus rates: helpfulness $76\%$,
env-awareness $80\%$, believability $72\%$, factual-accuracy $68\%$.
Average per-cell \sys{}-preference rate: $78.2\%$.

\paragraph{Inter-rater reliability.}
Pooled across all dimensions, Krippendorff's $\alpha_{\text{ord}}$ on
the raw 1--5 scores is $0.43$ (per-dimension range $0.34$--$0.47$),
indicating moderate agreement on absolute quality. Item-level
$\alpha$ on the preference difference $\Delta$ is much lower
($\alpha_{\text{ord}}{\in}[0.03, 0.17]$ per dimension) because raters
differ systematically in scale-use (one rater uses 5/1 polar ratings,
others use 4/2). The \emph{direction} of preference is nonetheless
highly consistent (above: $74\%$ \sys{}-consensus, $0\%$
Vanilla-consensus). We report all three numbers and treat the
pattern as: agreement on direction is strong, agreement on magnitude
is weak, and headline effect sizes pool through this magnitude
variance.

\paragraph{Sensitivity analysis.}
Excluding the one reverse-preference rater post-hoc yields $N{=}7$
and strengthens all four dimensions: helpfulness $\Delta{=}+1.85$,
$d_z{=}1.45$; environmental awareness $\Delta{=}+2.15$,
$d_z{=}1.62$; believability $\Delta{=}+1.86$, $d_z{=}1.41$; factual
accuracy $\Delta{=}+1.65$, $d_z{=}1.37$. We report this only as a
robustness check; the headline numbers in Table~\ref{tab:rq5} are
the $N{=}8$ primary analysis.

\paragraph{Per-category structure.}
The category ordering is consistent with the factual-grounding per-category
analysis: temporal ($+1.83$ avg) and memory ($+1.77$) at the top,
spatial ($+1.29$) and planning ($+1.37$) lower. Social ($+1.77$)
matches the factual-grounding finding that proactive probing helps most where
residual environmental uncertainty is concentrated.

\paragraph{Independent fabrication scan.}
Independently of the human ratings, we manually scanned all 50
\sys{} responses for fabricated proper-name entities
(locations or characters absent from the \textsc{AURATown} setup,
which has 20 named locations and 5 named characters). Two scenarios
contain explicit fabrications: scenario id 6 lists ``Bookstore, Art
Supply Store, Clothing Boutiques, Craft Stores, Gift Shops'' (none
exist among the 20 locations); scenario id 7 references ``Main
Street'' and ``The Cozy Corner Cafe'' (\textsc{AURATown} has neither
that street nor a cafe by that name; the only cafe is \emph{Sunrise
Cafe}). The other 48 responses cite only roster-real locations and
characters, giving a static-entity fabrication rate of $2/50 = 4\%$.
The scan is reproducible from the released response set against the
canonical roster in \texttt{demo/\allowbreak{}town/\allowbreak{}assets/\allowbreak{}town\_map.json}.

\paragraph{Methodological limitations.}
\textbf{(1) Sample size.} $N{=}8$ is small relative to managed
crowdsourcing studies. The consistency of direction across raters
and the large effect sizes mitigate but do not substitute for a
larger study; primary statistics should be read as evidence of a
large effect with substantial uncertainty around the precise
magnitude.
\textbf{(2) Recruitment.} Raters were recruited individually rather
than through a managed platform (Prolific, MTurk).
\textbf{(3) Dynamic-state factual accuracy is incompletely
measured.} The form did not display simulation scene state at query
time, so raters' factual\_accuracy judgments mix verifiable
static-entity checks (catchable from the on-form roster) with
trust-prior on dynamic-state claims (agent positions, current
activities). The independent fabrication scan above bounds the
static rate at $4\%$; precise quantification of dynamic-state error
requires deterministically recapturing scene state at each query's
generation time, which is future work.
\textbf{(4) IRR is moderate.} $\alpha \approx 0.4$ on raw scores
reflects scale-use heterogeneity across raters; we publish all
per-rater data so this variance is auditable.

\subsection{Probe-Budget Sweep (Full Multi-Seed)}
\label{sec:rq6}

\begin{figure}[t]
\centering
\resizebox{\columnwidth}{!}{%
\begin{tikzpicture}
\begin{axis}[
    width=0.92\columnwidth,
    height=5.0cm,
    xlabel={Probe budget $B$},
    ylabel={Grounding Accuracy (mean$\pm$std, 3 seeds)},
    axis y line*=left,
    ymin=0.84, ymax=0.89,
    xmin=-0.3, xmax=5.3,
    xtick={0,1,2,3,4,5},
    xlabel style={font=\footnotesize},
    ylabel style={font=\footnotesize, color=blue!70!black},
    xticklabel style={font=\footnotesize},
    yticklabel style={font=\footnotesize, color=blue!70!black},
    legend style={font=\footnotesize, at={(0.50,1.10)}, anchor=south, legend columns=2},
    grid=major,
    grid style={gray!20},
]
\addplot+[blue!70!black, thick, mark=*, mark size=2.5pt,
          error bars/.cd, y dir=both, y explicit]
coordinates {
    (0, 0.8621) +- (0, 0.0014)
    (1, 0.8743) +- (0, 0.0089)
    (2, 0.8621) +- (0, 0.0098)
    (3, 0.8628) +- (0, 0.0075)
    (4, 0.8577) +- (0, 0.0003)
    (5, 0.8681) +- (0, 0.0012)
};
\addlegendentry{GA}
\addplot[only marks, mark=o, mark size=6pt, draw=red!80, thick]
    coordinates {(0, 0.8621) (1, 0.8743)};
\addlegendentry{Pareto frontier}
\node[font=\tiny, anchor=south west, text=red!70] at (axis cs:1.05, 0.8755)
    {$B^{\star}{=}1$};
\end{axis}

\begin{axis}[
    width=0.95\columnwidth,
    height=5.0cm,
    ylabel={Latency per step (s)},
    axis y line*=right,
    axis x line=none,
    ymin=14, ymax=42,
    xmin=-0.3, xmax=5.3,
    ylabel style={font=\footnotesize, color=red!60!black},
    yticklabel style={font=\footnotesize, color=red!60!black},
    legend style={font=\footnotesize, at={(0.98,0.05)}, anchor=south east},
]
\addplot+[red!60!black, dashed, thick, mark=triangle*, mark size=2.5pt]
coordinates {
    (0, 17.3) (1, 21.4) (2, 25.8) (3, 31.1) (4, 35.3) (5, 39.3)
};
\addlegendentry{Latency}
\end{axis}
\end{tikzpicture}}
\caption{Probe-budget sweep, 3 seeds, 500 GA judgments per $(B, \text{seed})$ cell. Pareto frontier: $\{B{=}0, B{=}1\}$ (red rings); $B^{\star}{=}1$ adds $+0.012$ GA at $+4.1$\,s, every $B \ge 2$ is dominated. The same single probe buys $+13.7$\,pp implicit score in the social sub-regime (Table~\ref{tab:rq-tom-subcat}); routine GA is saturated.}
\label{fig:budget}
\end{figure}
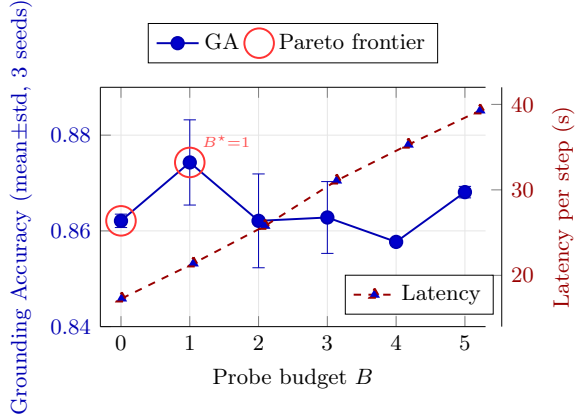

We vary the probe budget $B$ from 0 to 5 and measure GA and latency per step (500 GA judgments per budget per seed, averaged across 3 seeds $\{42, 123, 456\}$). Fig.~\ref{fig:budget} visualises the GA-vs-latency Pareto frontier.

\begin{table}[t]
\centering
\caption{Probe budget sweep, 3 seeds, 500 judgments per $B$ per seed. Pareto frontier: only $B \in \{0, 1\}$.}
\label{tab:budget}
\small
\resizebox{\columnwidth}{!}{%
\begin{tabular}{cccc}
\toprule
\textbf{Budget $B$} & \textbf{GA (mean $\pm$ std)} & \textbf{Latency (s)} & \textbf{Pareto?} \\
\midrule
0 & $0.8621 \pm 0.0014$ & $17.3 \pm 0.8$ & \checkmark \\
1 & $\mathbf{0.8743 \pm 0.0089}$ & $21.4 \pm 0.8$ & \checkmark \\
2 & $0.8621 \pm 0.0098$ & $25.8 \pm 1.4$ & \\
3 & $0.8628 \pm 0.0075$ & $31.1 \pm 1.4$ & \\
4 & $0.8577 \pm 0.0003$ & $35.3 \pm 1.4$ & \\
5 & $0.8681 \pm 0.0012$ & $39.3 \pm 1.8$ & \\
\bottomrule
\end{tabular}%
}
\end{table}

The Pareto frontier contains exactly two points, $\{B{=}0, B{=}1\}$; $B{=}1$ is the peak. $B{=}0 \to B{=}1$ adds $+0.0122$ GA ($+1.4\%$ relative) at $+4.1$\,s latency; every $B \ge 2$ is Pareto-dominated. An earlier single-seed pilot reported $B^* {=} 2$; the 3-seed replication withdraws that claim. On the routine-grounding metric the curve is consistent with a monotonically saturating information channel rather than a clean crossover: one probe fills the scene snapshot, subsequent probes contribute little. The same single probe buys $+13.7$\,pp implicit score in the social sub-regime (Section~\ref{sec:rq-tom}, Table~\ref{tab:rq-tom-subcat}), confirming that the budget-sweep's small absolute gain is a feature of the metric, not the mechanism.

\subsection{Judge Disagreement}
\label{sec:appendix-judge}

We quantify judge disagreement between the rule-based pre-filter and the LLM judge across 7{,}500 multi-seed judgments in Table~\ref{tab:judge_disagree}. Disagreement varies by dimension: time appropriateness shows the highest rate ($61.9\%$, rule stricter in $99.5\%$), location consistency $48.4\%$ ($92.3\%$ rule-stricter), plan adherence $32.0\%$ ($83.0\%$ LLM-stricter), social awareness $13.5\%$ ($98.8\%$ LLM-stricter), memory utilisation $0.0\%$. We resolve disagreements by taking the stricter score (conservative grounding estimate).

\begin{table}[t]
\centering
\caption{Rule-based vs.\ LLM judge disagreement across 7{,}500 grounding judgments (3 seeds $\times$ 5 conditions $\times$ 500 per condition).}
\label{tab:judge_disagree}
\small
\resizebox{\columnwidth}{!}{%
\begin{tabular}{lcccc}
\toprule
\textbf{Dimension} & \textbf{Disagree} & \textbf{Rule Stricter} & \textbf{LLM Stricter} \\
\midrule
Time Appropriateness & $61.9\%$ & $99.5\%$ & $0.5\%$ \\
Location Consistency & $48.4\%$ & $92.3\%$ & $7.7\%$ \\
Plan Adherence       & $32.0\%$ & $17.0\%$ & $83.0\%$ \\
Social Awareness     & $13.5\%$ & $1.2\%$  & $98.8\%$ \\
Memory Utilisation   & $0.0\%$  & ---      & ---      \\
\bottomrule
\end{tabular}%
}
\end{table}

\section{Second Human-Evaluation Arm: Intent vs.\ NoIntent}
\label{sec:rq5-nointent}

The N=8 study in Appendix~\ref{sec:rq5} compares \sys{} against Vanilla and therefore cannot speak to the Intent-vs.-NoIntent contrast. Reviewers correctly identified this gap, and we ran a second arm.

\paragraph{Design.} 50 items sampled from the 100-query benchmark, stratified 10 per subcategory, pairing the seed-42 \texttt{tom} and \texttt{no\_intent} answers already logged for Table~\ref{tab:rq-intent-v2-main} (no new generation). Presentation order randomised per item (\sys{} appears as response A in 30 of 50). The \texttt{[heads-up]} prefix was stripped from \sys{} answers, which would otherwise identify the condition. Unlike the first arm, each item displays the scene's ground-truth public \emph{and} hidden state, since factual accuracy is not judgeable from a response pair alone; the benchmark's \texttt{implicit\_need} annotation is withheld, as it is the answer key. Eight volunteer raters, distinct from the authors, each scored all 50 items on the same four dimensions (1--5). Mean pairwise inter-rater correlation is $r{=}0.24$, lower than the first arm, as expected when both systems are competent and differences are subtle.

\paragraph{Result.} The arm does not show a human-perceptible advantage for \sys{}-Intent (Table~\ref{tab:rq5-nointent}). Three dimensions favour \sys{} directionally but none reaches significance; on factual accuracy \sys{} rates significantly \emph{lower} ($-0.29$, $p{=}0.008$, rater-aggregated Wilcoxon), with 0 of 8 raters placing \sys{} higher.

\begin{table}[!ht]
\centering
\caption{Second human-evaluation arm: \sys{}-Intent vs.\ NoIntent, rater-aggregated means over 8 raters $\times$ 50 items. $p$: rater-aggregated Wilcoxon.}
\label{tab:rq5-nointent}
\small
\setlength{\tabcolsep}{4pt}
\resizebox{\columnwidth}{!}{%
\begin{tabular}{lcccc}
\toprule
\textbf{Dimension} & \textbf{\sys{}} & \textbf{NoIntent} & $\Delta$ & $p$ \\
\midrule
Helpfulness   & $3.49$ & $3.37$ & $+0.12$ & $0.48$ \\
Env.\ awareness & $3.72$ & $3.50$ & $+0.22$ & $0.08$ \\
Factual accuracy & $3.65$ & $3.93$ & $-0.29$ & $\mathbf{0.008}$ \\
Believability & $3.65$ & $3.63$ & $+0.02$ & $0.81$ \\
\bottomrule
\end{tabular}%
}
\end{table}

\paragraph{Reading.}
Three things bound this result. The arm rated the full \texttt{IntentFrame} condition, which our own component ablation (Table~\ref{tab:component-ablation}) shows is dominated by gap-derived budget routing alone; the variant we now recommend was not the variant rated. \sys{}-Intent's answers also run nearly twice as long as NoIntent's, and rater comments penalise unrequested elaboration rather than the surfacing of the implicit need itself. Most importantly, implicit-need coverage and general response quality are different properties: the first is scored against a stated hidden need, the second is a holistic preference, and only the first is what \sys{} claims. We report the arm because we committed to it, and read it as evidence that the \texttt{recommended\_probes} half of the frame costs perceived answer quality, which is consistent with the ablation.

\section{Heuristic vs.\ LLM IntentInferrer (Backend Ablation)}
\label{sec:appendix-heuristic}

The \texttt{IntentFrame} pipeline of Section~\ref{sec:method-intent}
admits two backends: a deterministic \texttt{HeuristicIntentInferrer}
(rule-based gap estimation from a small vocabulary of social/private-state
markers) and \texttt{LLMIntentInferrer} (gpt-4o-mini producing the
frame as JSON). The main implicit-intent run uses the LLM backend. We
isolate the inferrer's contribution by running the SAME 25 implicit-intent queries
$\times$ 3 seeds with the heuristic backend; all other plumbing (gap-to-budget map,
directed probe loop, judge) is identical.

\begin{table}[t]
\centering
\caption{Per-subcategory implicit score: heuristic-backend vs.\ LLM-backend
\sys{} Intent (25 queries $\times$ 3 seeds, $N{=}15$ per cell). The heuristic
collapses to $\approx$ Literal performance on \emph{availability} and
\emph{latent\_goal} because its surface-keyword gap estimator does not
fire on lexically-decoupled queries; the LLM backend recovers the gap
and routes the appropriate probes.}
\label{tab:heuristic-vs-llm-intent}
\small
\resizebox{\columnwidth}{!}{%
\begin{tabular}{lcccc}
\toprule
\textbf{Subcategory} & \textbf{Literal} & \textbf{Heuristic Intent} & \textbf{LLM Intent} & \textbf{LLM gain over heuristic} \\
\midrule
availability     & $0.29$ & $0.27$ & $\mathbf{0.67}$ & $+0.40$ \\
mood             & $0.29$ & $0.33$ & $\mathbf{0.84}$ & $+0.51$ \\
appropriateness  & $0.40$ & $0.47$ & $\mathbf{0.84}$ & $+0.37$ \\
latent\_goal     & $0.00$ & $0.00$ & $\mathbf{0.75}$ & $+0.75$ \\
second\_order    & $0.09$ & $0.77$ & $\mathbf{0.92}$ & $+0.15$ \\
\midrule
\textbf{Overall ($N{=}75$)} & $0.216$ & $0.368$ & $\mathbf{0.803}$ & $+0.44$ \\
\midrule
probes (mean)    & $0.0$ & $0.32$ & $1.48$ &  \\
latency (s)      & $1.3$ & $1.9$ & $13.8$ &  \\
\bottomrule
\end{tabular}%
}
\end{table}

\paragraph{Reading.}
The heuristic recovers $+0.18$ over Literal on aggregate but lags the
clean calibrated LLM backend by $-0.44$. Per-subcategory: the heuristic
stays near Literal on availability ($0.27$ vs.\ $0.29$) and ties Literal
on latent\_goal ($0.00$ vs.\ $0.00$)
because its trigger vocabulary (``available'', ``mood'', ``appropriate'',
``up to'') does not match the surface of those query classes
(``where is X?'', ``what is X up to?''); the gap estimator returns $0$
and the heuristic falls through to the literal answer. On
second\_order it does well ($0.77$) because ``thinks'', ``believes'',
and ``perspective'' do appear in the trigger set. The contrast
quantifies what the LLM backend buys: $+0.44$ aggregate at $7.3\times$
the heuristic latency, with the LLM doing the work precisely where
surface cues fail.

\paragraph{Implication.}
Section~\ref{sec:rq-tom}'s headline gain is attributable to the LLM
backend, not the gap-to-budget mapping or the probe-loop machinery.
Distilling a faster intent classifier from the LLM backend is a
practical optimisation; the rule-based shortcut is not.

\paragraph{The fixed probe order used by the recommended configuration.}
The deployment configuration recommended in Section~\ref{sec:rq-tom} pairs the gap-derived ceiling with a deterministic, intent-agnostic tool order. Second-order queries probe \texttt{get\_agent\_belief\_about} for the named believer and target. Every other query walks \texttt{get\_agent\_private\_state} over the agent the query names, or, when it names none, over the agents at the scene's focus location, truncated to the ceiling. Nothing in this order consults the \texttt{IntentFrame} beyond the ceiling itself, which is what makes it a clean control for the shortlist.

\section{Cross-Domain Sanity Checks}
\label{sec:appendix-crossdomain}

The main-body results live in the social-simulation regime where the
\envagent{} pipeline was designed. To probe how the architecture
transfers, we ran four additional benchmarks outside that regime; we
report them here as sanity checks rather than as headline
contributions. \textbf{All four runs are single-seed} (vs.\ 3 seeds for
the main implicit-intent and factual-grounding experiments) and per-run
sample sizes vary: InteractiveBench Puzzle $20$ episodes, Trust
$6$ games/condition, Math $20$ questions, MemoryArena $n{=}1$ paper
across $5$ subtasks. Trust ($n{=}6$) and MemoryArena ($n{=}1$) are
underpowered for statistical inference and are reported as
observational anchors only; we do not run paired tests on them.

\begin{figure}[!htbp]
\centering
\includegraphics[width=\columnwidth]{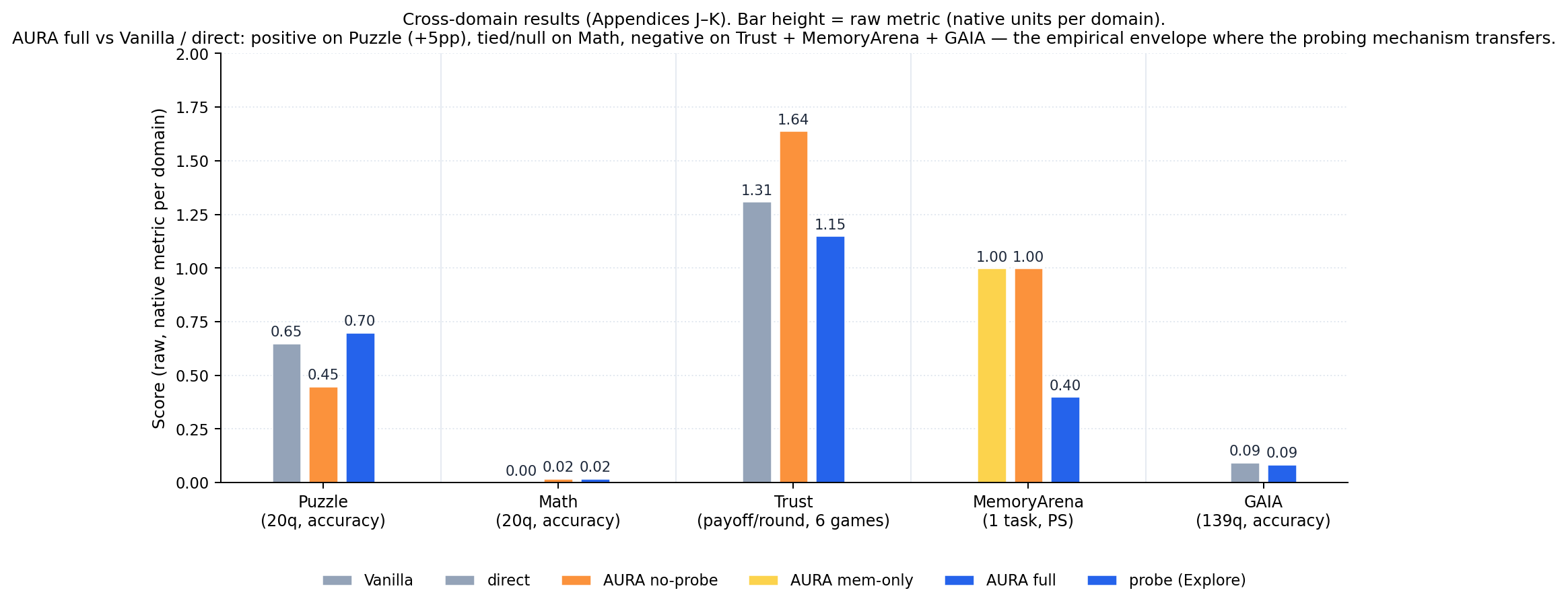}
\caption{Cross-domain results (\textbf{all single-seed; sample sizes per panel vary, see text}). Bar height is the raw metric in each domain's native units (accuracy, payoff/round, partial-success score). \sys{} full beats Vanilla on Puzzle ($+5$\,pp absolute, $n{=}20$ episodes), ties on Math ($n{=}20$, judge-saturated), and underperforms on Trust ($n{=}6$ games/cond.\ --- underpowered, descriptive only), MemoryArena ($n{=}1$ paper $\times$ 5 subtasks --- anecdotal), and GAIA ($n{=}139$, see Appendix~\ref{sec:appendix-gaia}). Together with the scope condition in Section~\ref{sec:rq2}, these data points sketch the empirical envelope of the probing mechanism: it transfers when residual environmental uncertainty after passive perception is non-trivial \emph{and} tool returns are structurally extractable.}
\label{fig:cross-domain}
\end{figure}

\paragraph{InteractiveBench --- Puzzle (text adventure).}
20 multi-turn puzzle episodes per condition, single seed, gpt-4o-mini.
\sys{} (Full) reaches $0.70$ accuracy, Vanilla LLM $0.65$, \sys{}
(No-Probe) $0.45$. The architecture-without-probing
under-performs Vanilla here because the No-Probe pipeline injects an
empty environment context that consumes turns; restoring probing
recovers and slightly exceeds Vanilla ($+5$\,pp absolute, $+7.7\%$
relative). This is the cleanest cross-domain signal we observe.

\paragraph{InteractiveBench --- Trust (iterated prisoner's dilemma).}
6 games per condition against three opponent strategies (TFT, GrimTrigger, Random). Aggregate payoff per round: Vanilla $1.31$, \sys{}
No-Probe $1.64$, \sys{} Full $1.15$. The aggregate is dominated by
opponent-mix imbalance: against deterministic strategies all conditions reach the cooperative equilibrium of
$2.00$ payoff/round; against the noisy random opponent, the
sample sizes differ ($n_{\text{rounds}} = 17, 16, 28$) and the
\sys{} Full mean is $0.96$. We interpret this as the negative
counterpart to Puzzle: when the ``environment'' is an adversarial
opponent whose private state cannot be probed, additional context
gathering does not help and may slow convergence. We do not claim a
positive result on Trust.

\paragraph{InteractiveBench --- Math (LaTeX-conditional QA).}
20 questions per condition, single seed; all three conditions reach
$\approx 1.9\%$. The LLM judge is conservative on \LaTeX{}-formatted
answers and returns repeated parse errors (e.g., ``Invalid
\textbackslash{}escape'') that we manually verified mark correct
agent responses as incorrect. We report the result as null and
attribute it to judge brittleness, not to a meaningful gap between
conditions; a per-domain custom judge is future work.

\paragraph{MemoryArena (research-paper Q\&A, formal reasoning) --- \textbf{anecdotal $n{=}1$}.}
We ran a single-paper smoke test (1 paper $\times$ 5 subtasks, single seed), which is too small to support a quantitative claim and we do not run statistical tests on it. We report the observation for completeness: \sys{} (memory only)
and \sys{} (no\_probe) reach $5/5$ partial-success ($PS{=}1.0$);
\sys{} (full) reaches $2/5$ ($PS{=}0.4$). The full-pipeline
under-performs because probe results pollute the long mathematical
context with environment-style snippets---the same
``hallucinated context'' failure mode observed at $B \ge 3$ (probes add noise faster than they reduce errors), here amplified by the long-context
formal-reasoning regime. Sample is too small to draw a quantitative
claim, but the direction is consistent with the scope condition: the
probing mechanism degrades when the task is symbolic and the tool
returns are off-distribution.

\paragraph{Cross-domain takeaway.}
The architecture transfers positively to a multi-turn puzzle domain,
fails to transfer to adversarial games where opponent state cannot be
probed, is masked by judge brittleness on \LaTeX{} math, and
under-performs on long-context formal reasoning where probe outputs
pollute the context. Together with the GAIA negative-transfer result
in Appendix~\ref{sec:appendix-gaia}, these data points describe the
empirical envelope of the probing mechanism beyond the primary
social-simulation setting.

\section{FANToM External ToM Benchmark (Null Result)}
\label{sec:appendix-fantom}

To probe whether AURA's intent-directed mechanism transfers to a standard, externally-validated theory-of-mind benchmark, we ran a 400-question stratified split of FANToM~\citep{kim2023fantom}: $80$ questions per type across the five FANToM question types --- \emph{beliefQA} (multi-choice belief), \emph{answ\_bin}, \emph{answ\_list} (answerability binary / list), \emph{info\_bin}, \emph{info\_list} (info-accessibility binary / list). The $80$ \emph{beliefQA} questions are sub-stratified $20{+}20{+}20{+}20$ across first-/second-order $\times$ accessible/inaccessible. Backbone is \texttt{gpt-4o-mini}, seed=$42$ (the OpenAI \texttt{seed} kwarg is accepted in this run, so backbone draws are deterministic when supported). Source: \texttt{\allowbreak{}\-run\_fantom\_full.py}; data: \texttt{\allowbreak{}evaluation/\allowbreak{}\-results/\allowbreak{}fantom\_full\_seed42.json}.

\begin{table}[t]
\centering
\caption{FANToM accuracy by question type ($n{=}80$ each, $400$ total). Three conditions on the same backbone: \emph{Literal} = single-call, no tools; \emph{NoIntent} = ReAct-style tool use up to budget 3; \emph{Intent} = full AURA pipeline with IntentInferrer. Last column: per-question paired contrasts.}
\label{tab:fantom}
\small
\resizebox{\columnwidth}{!}{%
\begin{tabular}{lcccl}
\toprule
\textbf{Question type} & \textbf{Literal} & \textbf{NoIntent} & \textbf{Intent} & \textbf{Notes} \\
\midrule
beliefQA   & $\mathbf{0.675}$ & $0.662$ & $0.662$ & FO/SO $\times$ acc/inacc balanced \\
answ\_bin  & $\mathbf{0.725}$ & $0.525$ & $0.525$ & Literal wins \\
answ\_list & $0.375$ & $0.362$ & $\mathbf{0.412}$ & Intent wins (small) \\
info\_bin  & $\mathbf{0.887}$ & $0.863$ & $0.875$ & Literal wins (small) \\
info\_list & $0.425$ & $0.562$ & $\mathbf{0.625}$ & Intent wins \\
\midrule
\textbf{Overall} & $0.617$ & $0.595$ & $\mathbf{0.620}$ & --- \\
\bottomrule
\end{tabular}%
}
\end{table}

\paragraph{Statistical tests (per-question paired, $n{=}400$).}
Intent vs.\ Literal: $\Delta\text{acc}{=}{+}0.003$, paired $t$ $p{=}0.92$; McNemar $p{=}1.0$. Intent vs.\ NoIntent: $\Delta\text{acc}{=}{+}0.025$, paired $t$ $p{=}0.26$; McNemar $p{=}0.31$. Both contrasts null.

\paragraph{Interpretation.}
FANToM's narrative ToM questions ship the full conversation in-context, so a competent backbone can answer literally without retrieval; the IntentFrame's gap calibration finds little to direct, and the additional probe budget adds latency ($6.5$\,s vs.\ $0.9$\,s for Literal) without lifting accuracy. FANToM is therefore a clean negative case for AURA's mechanism --- the residual uncertainty $H(\mathcal{E}\mid\mathcal{B})$ after passive context assembly is already low, so probing has nothing to reduce. This bounds the probing claim's scope: AURA helps when private state lies behind a tool-mediated information frontier (multi-agent simulation, scene-grounded queries with hidden private state), and does not help on narrative ToM transcripts already in-context.

\paragraph{Backbone-capability corroboration.}
The Intent condition's IntentInferrer fell back to the deterministic heuristic $0/400$ times on \texttt{gpt-4o-mini} (vs.\ $23/25$ on \texttt{gemini-2.5-flash} in Section~\ref{sec:rq-tom}). The contrast confirms that the cross-backbone Gemini regression in Table~\ref{tab:rq-tom-backbone} is a backbone-capability failure (Gemini's structured-JSON adherence) rather than a problem with the IntentInferrer's prompt or the AURA pipeline.

\section{LoCoMo Long-Term Conversational Memory (Partial Transfer)}
\label{sec:appendix-locomo}

To probe whether the AURA pipeline transfers to the long-term-conversational-memory recall regime, we ran a 200-question stratified split of LoCoMo~\citep{maharana2024locomo} (10 conversations $\times$ 19--32 sessions $\times$ ${\sim}20$ turns/session, $2{,}206$ QA total across 5 categories). Backbone \texttt{gpt-4o-mini}, seed=$42$, 8 parallel workers, OpenAI \texttt{seed} kwarg accepted. Source: \texttt{\allowbreak{}\-run\_locomo\_smoke.py} (re-runnable at \texttt{LOCOMO\_N=200}); data: \texttt{\allowbreak{}evaluation/\allowbreak{}\-results/\allowbreak{}locomo\_smoke.json}. Adapter at \texttt{\allowbreak{}locomo\_eval.py} maps a conversation's session list to AURA's \texttt{(scene, memories, query, available\_tools)} quadruple, with three simulated probe tools (\texttt{get\_session(n)}, \texttt{search\_by\_speaker}, \texttt{list\_sessions\_on\_date}). Scoring follows the LoCoMo paper exactly: token-F1 with stem normalisation, multi-answer F1 for cat-1, semicolon-alternative for cat-3.

\begin{table}[t]
\centering
\caption{LoCoMo 200-question result (1 seed, gpt-4o-mini). \emph{Literal} = single LLM call with session catalog only; \emph{NoIntent} = ReAct-style tool use; \emph{Intent} = full AURA pipeline with IntentInferrer. F1 reported because token-EM is near-zero across all conditions (LoCoMo gold answers are dense phrases that natural prose rarely matches set-equal --- same as the upstream paper). \emph{Fallback} = LLMIntentInferrer parse failures.}
\label{tab:locomo}
\small
\resizebox{\columnwidth}{!}{%
\begin{tabular}{lccccc}
\toprule
\textbf{Condition} & \textbf{F1} & \textbf{EM} & \textbf{Mean lat (s)} & \textbf{Probes (mean)} & \textbf{Fallback} \\
\midrule
Literal   & $0.042$ & $0.020$ & $1.16$ & $0.00$ & --- \\
NoIntent  & $0.173$ & $0.025$ & $2.91$ & $1.33$ & $0/200$ \\
\sys{} Intent & $\mathbf{0.192}$ & $0.025$ & $6.19$ & $1.42$ & $0/200$ \\
\bottomrule
\end{tabular}%
}
\end{table}

\paragraph{Statistical tests (per-question paired, $n{=}200$).}
Intent vs.\ Literal: $\Delta\text{F1}{=}{+}0.151$, paired $t{=}8.11$, $\mathbf{p<10^{-15}}$ --- highly significant.
Intent vs.\ NoIntent: $\Delta\text{F1}{=}{+}0.020$, paired $t{=}1.09$, $p{=}0.28$ --- \textbf{not significant}.

\paragraph{Interpretation.}
The architecture-and-tools layer (NoIntent: probe-tool harness with no intent reframing) carries most of the gain from $0.042$ to $0.173$ F1; AURA's IntentInferrer adds only an additional $+0.020$ F1, which is not statistically significant on this sample. Read together with the FANToM null (Appendix~\ref{sec:appendix-fantom}), LoCoMo gives the same picture: the architectural pipeline transfers to long-term-recall and to narrative ToM, but the intent-direction stage's marginal contribution \emph{above} a tool-using baseline is regime-specific. It is significant on AURATown's hand-designed implicit-intent set (Section~\ref{sec:rq-tom}) where private-state needs are deliberately hidden behind the surface form, and small-or-null on benchmarks where the underlying QA does not require this kind of lexical-vs-implicit gap inference.

The $0/200$ \texttt{IntentInferrer} fallback rate (alongside $0/400$ on FANToM) is the cleanest external corroboration that the original \texttt{gemini-2.5-flash} cross-backbone regression (Section~\ref{sec:rq-tom}) is a backbone format-compliance failure on Gemini's part, not a problem with AURA's prompt or schema.

\section{GAIA Cross-Domain Run (Negative Transfer)}
\label{sec:appendix-gaia}

We ran the GAIA Level-1/2 question set (139 questions per condition,
seed 42, gpt-4o-mini) under two conditions: \emph{probe} (the \sys{}
Explore stage with the OpenAI Responses API \texttt{web\_search}
tool) and \emph{direct} (single-pass LLM, no tools).

\begin{table}[t]
\centering
\caption{GAIA Level-1/2 (seed 42, $n{=}139$ per condition).}
\label{tab:gaia}
\small
\resizebox{\columnwidth}{!}{%
\begin{tabular}{lcccc}
\toprule
\textbf{Condition} & \textbf{Acc.} & \textbf{Lat.\,(s)} & \textbf{Avg.\ tool calls} & \textbf{Acc.\ L1 / L2} \\
\midrule
direct (no probe) & $\mathbf{0.094}$ & $0.92$  & $0.0$ & $0.094 / 0.093$ \\
probe (Explore on)   & $0.086$          & $20.5$  & $3.5$ & $0.075 / 0.093$ \\
\bottomrule
\end{tabular}%
}
\end{table}

\paragraph{Reading.}
On Level-1 (single-step factual lookup) probing \emph{degrades}
accuracy by $-1.9$\,pp absolute; on Level-2 (multi-step reasoning)
the two conditions tie. Total cost: probing spends $22\times$ more
wall time and $3.5$ tool calls per question for no aggregate benefit.

\paragraph{Why the transfer fails.}
GAIA's environment is a search engine wrapped in an LLM-simulated
browser. Each ``probe'' is itself an LLM call summarising a web page,
not a structured-state read against a ground-truth simulator as in
\textsc{AURATown}. Two consequences: (i) probe outputs inherit the
backbone's failure modes (hallucinated facts compound across the
probe loop), and (ii) the residual $H(\mathcal{E} \mid \mathcal{B})$
that probing was designed to reduce on social-availability queries is
not the bottleneck on web-grounded factoid questions, where the
backbone's parametric knowledge dominates. This is the boundary case
of the scope condition stated in
Section~\ref{sec:rq2}: probing helps when (a) residual uncertainty
after passive perception is non-trivial \emph{and} (b) tool returns
are structurally extractable. GAIA fails (b).

\paragraph{What we report.}
We report GAIA as evidence about a regime in which the bounded-probing
mechanism does not transfer. Coupling this with the InteractiveBench
Trust negative and the MemoryArena formal-reasoning degradation, the
empirical envelope of \sys{}'s probing contribution is structured
social environments with extractable, non-hallucinated state probes.

\section{Memory and Enrichment Protocol Details}
\label{sec:appendix-memory}

\paragraph{Weighted retrieval.}
For each memory $m$, query $q$, current time $t$:
\begin{align}
    \text{score}(m, q, t) &= w_r \cdot \exp(-\lambda (t - m.t)) \notag\\
    &\quad + w_p \cdot \tfrac{m.\text{importance}}{10} \notag\\
    &\quad + w_v \cdot \text{sim}(q, m.c)
    \label{eq:memory_score}
\end{align}
Default weights $(w_r, w_p, w_v) = (0.3, 0.4, 0.3)$, decay $\lambda = 0.01$. $\text{sim}(\cdot)$ is a keyword-based relevance function in the current prototype.

\paragraph{Memory types.}
Following \citet{tulving1972episodic}: \emph{Observation} (direct percepts), \emph{Conversation} (dialogues), \emph{Reflection} (synthesised insights every $\theta = 10$ observations), \emph{Plan} (daily schedules and goals).

\paragraph{Importance scoring.}
When an LLM endpoint is available, importance is scored via a dedicated prompt (``Rate the importance of this event 1--10''); otherwise a keyword-based heuristic assigns scores based on emotional salience.

\paragraph{Three-stage enrichment protocol.}
User query $q$ to agent $a_i$ triggers: (1) context gathering (perception + recent memories), (2) proactive probing (1-step for latency), (3) enriched generation with the structured context prepended. Enrichment metadata (tools called, context gathered) is exposed to the user for interpretability.

\section{Reproducibility Checklist}
\label{sec:appendix-reproducibility}

\paragraph{Code and data.}
All experiments are driven by the runners in \texttt{scripts/} and \texttt{run\_experiments.py}, with results saved to \texttt{evaluation/results/}. Each table or figure traces to a specific result JSON via \texttt{evaluation/results/MANIFEST.md}; the manifest covers factual-grounding multi-seed, factual-grounding Fixed-Probe and GapRouted Pareto controls, the privacy-sensitive distractor slice, implicit-intent clean-prompt multi-seed, fixed-private/oracle-intent controls, no-few-shot prompt ablation, Plan-and-Solve and Reflexion adapter outputs, FANToM 400q, LoCoMo 200q, the four cross-backbone runs, implicit-intent IAA returns, and the strict-precision rescore. The public release at \url{https://github.com/innovation64/AURA} includes the \sys{} agent library, the \textsc{AURATown} simulation, all 50 environment-grounded queries with their templates, all 100 implicit-intent queries (4 scenes $\times$ 25) with subcategory and target labels, and all per-condition / per-seed details that back the numbers reported here; the implicit-intent benchmark is additionally mirrored on the HuggingFace Hub as \texttt{innovation64/aura-implicit-intent} (CC~BY~4.0). The component-ablation and gap-calibration analyses (Section~\ref{sec:rq-tom}) are regenerated by \texttt{\allowbreak{}\-gap\_calibration\_analysis.py} from \texttt{rq\_intent\_v2\_multiseed.json} and by the \texttt{budget\_only} condition of the v2 runner (\texttt{rq\_intent\_v2\_budget\_only\_multiseed.json}).

\paragraph{Models and APIs.}
Agent backbone and judge are both \texttt{gpt-4o-mini} via the OpenAI Chat Completions API. Cross-backbone tests additionally use \texttt{claude-haiku-4-5} (Anthropic), \texttt{qwen-plus} (Alibaba), and \texttt{gemini-2.5-flash} (Google), each via its vendor's official API. Backbone temperature is 0.7 for action decisions, 0.1 for the judge and the IntentInferrer; these values govern \textsc{AURATown} rollouts. The implicit-intent benchmark (\S\ref{sec:rq-tom}) is a separate pipeline in which both answer generation and judging run at temperature 0.1. We pass the \texttt{seed} parameter to the OpenAI Chat Completions API where the SDK accepts it; on rejection, the LLM engine falls back to a no-seed call and flips an internal flag for the rest of the run. Cross-backbone runs do not all support the \texttt{seed} kwarg, so the multi-seed claim is bounded as: ``Python-level random sampling and OpenAI \texttt{seed} when supported.''

\paragraph{Determinism caveats.}
Multi-seed paired tests for factual grounding / component ablation / implicit-intent vary the seed both at Python's \texttt{random} module (controlling query ordering, agent shuffles, and template instantiation) and at the OpenAI API \texttt{seed} kwarg. Stochastic backbone decoding above the seed kwarg may still introduce per-seed variance; this is acknowledged in the Limitations and is consistent with the OpenAI API documentation that \texttt{seed} provides a best-effort, not strict, determinism guarantee. The simulation server itself derives its world generation from \texttt{TownConfig.world\_seed} and the agent backbone derives its outputs from \texttt{TownConfig.llm\_seed}; both are propagated through \texttt{/api/reset}.

\paragraph{Strict-paired factual grounding.}
The factual-grounding runner (\texttt{\allowbreak{}run\_experiments.py:run\_rq2}) operates in two phases. Phase A: reset with the experiment seed, warm 10 steps, then advance one tick per query position to capture 50 ground-truth snapshots. Phase B: each AURA condition resets with the same seed, warms 10, and replays the snapshots; chats run with \texttt{read\_only=True} so they do not write event log or memory, leaving the per-tick trajectory deterministic across conditions. External baselines (Vanilla, Static, ReAct, Reflexion, Plan-and-Solve) receive the same Phase-A snapshots. Output JSON stamps a \texttt{\_paired\_snapshot\_meta} block recording \texttt{seed}, \texttt{n\_snapshots}, and \texttt{shared\_\allowbreak{}snapshots\_\allowbreak{}across\_\allowbreak{}conditions=true}.

\paragraph{Aggregate scripts.}
\texttt{\allowbreak{}\-rescore\_rq2\_strict.py} recomputes the strict precision rescore (Appendix~\ref{sec:appendix-strict-rescore}) from the multi-seed factual-grounding details. \texttt{\allowbreak{}\-aggregate\_rq2\_multiseed.py} collapses per-seed details into the canonical multi-seed summary. \texttt{\allowbreak{}\-run\_rq2\_fixed\_probe.py} and \texttt{\allowbreak{}\-run\_rq2\_aura\_gap\_routed.py} regenerate the saturated and gap-routed factual-grounding controls; \texttt{\allowbreak{}\-rq2\_pareto\_analysis.py} computes FA/probe/disclosure contrasts and \texttt{\allowbreak{}\-plot\_rq2\_pareto.py} renders Figure~\ref{fig:rq2-pareto}. \texttt{\allowbreak{}\-run\_privacy\_distractor.py} regenerates the 30-query forbidden-tool slice. \texttt{\allowbreak{}\-run\_implicit\_intent\_full.py} and \texttt{\allowbreak{}\-run\_implicit\_intent\_v2.py} regenerate the primary 25-query and expanded 100-query (4-scene) implicit-intent runs respectively; the v2 runner supports \texttt{--resume} for interrupted multi-seed jobs and a \texttt{--validate-only} pre-flight pass on the benchmark JSON. \texttt{\allowbreak{}\-compute\_irr.py} produces the IAA Cohen's $\kappa$ on the implicit-intent subcategory labels. \texttt{\allowbreak{}\-rq5\_rater\_aggregated.py} produces the per-rater aggregated Wilcoxon used in Table~\ref{tab:rq5}. \texttt{\allowbreak{}\-run\_fantom\_full.py} regenerates the FANToM 400-question external bench. \texttt{\allowbreak{}\-run\_locomo\_smoke.py} regenerates the LoCoMo run. \texttt{\allowbreak{}\-audit\_paper\_numbers.py} re-derives every cited headline from the JSON files and is the recommended pre-submission check.

\paragraph{Hardware.}
All experiments run on a single CPU machine; the LLM is invoked via remote API. No GPU is required. Wall-clock budgets per run: factual-grounding multi-seed ${\sim}30$\,min/seed; implicit-intent ${\sim}10$\,min/seed; FANToM 400 questions ${\sim}9$\,min; LoCoMo 200 questions ${\sim}6$\,min; PnS-on-implicit-intent multi-seed ${\sim}16$\,min total; Reflexion + PnS on factual grounding multi-seed combined ${\sim}2.5$\,h. Total API spend for all reported experiments under \$15 USD.

\section{Ethics Statement}
\label{sec:appendix-ethics}

\paragraph{Research artefacts.}
This paper studies an LLM agent architecture and accompanying social simulation. Human involvement consists of the 8 voluntary annotators of the first helpfulness study (\sys{} vs.\ Vanilla), 8 voluntary raters of the second arm (\sys{}-Intent vs.\ NoIntent, Appendix~\ref{sec:rq5-nointent}), and 2 additional annotators for the implicit-intent label audit. Annotators were friends and colleagues of the authors who consented to evaluate anonymous response pairs or query labels; no demographic data was collected; no personally identifying information appears in saved annotations. Annotators received no compensation; each task was a one-time, optional 15--30 minute evaluation. We retain a dissenting rater in the primary helpfulness analysis to avoid post-hoc exclusion bias and we report rater-level transparency (per-rater preference rates, Krippendorff's $\alpha$) so that readers can inspect inter-annotator dynamics.

\paragraph{Simulation content.}
\textsc{AURATown}'s 5 agents have hand-authored profiles (occupation, personality, daily routine). The names are fictional. Agent profiles deliberately span ages 20--68 and include contrasting personality types but are not intended to represent any real demographic distribution. The simulation does not emulate real people, real businesses, or real locations.

\paragraph{Misuse considerations.}
The IntentFrame mechanism produces a structured estimate of a user's implicit information need from a surface query. In a benign deployment this is used to surface relevant context the user did not lexically request; in an adversarial deployment, the same machinery could be used to surface information the user has not consented to share. The Explore stage's tool whitelist is the principal mitigation: tools are scoped to the simulation's structured environment state, not to external data sources. Practitioners deploying a similar IntentInferrer over real personal-data tools should constrain the whitelist accordingly and surface the inferred implicit need to the user (the heads-up prefix in our \texttt{Interact} stage is a minimal version of this).

\paragraph{Energy and compute.}
All experiments run on remote API endpoints; the only local compute is a single CPU process running the AURATown simulation server. We did not benchmark API energy use; total token consumption across all reported runs is approximately 8 million input tokens and 200 thousand output tokens against \texttt{gpt-4o-mini}.

\paragraph{Limitations as ethics.}
The paper's empirical claims are scoped to the regime where structured environment access matters (single-user situated queries with hidden private state). Where the mechanism does not transfer ( narrative ToM in FANToM, web-grounded factoid in GAIA, formal reasoning in MemoryArena ) we report negative results in Appendix~\ref{sec:appendix-fantom}, \ref{sec:appendix-gaia}, and \ref{sec:appendix-crossdomain}. We avoid abstracting from these specific findings to general claims about LLM theory-of-mind or proactive assistance.

\section{Collective Behaviour and Supplementary Discussion}
\label{sec:appendix-extra}


\subsection{Collective Behaviour in Multi-Agent \sys{}Town}
\label{sec:collective-tom}

In the 200-step SOTOPIA run, 5 agents with independent \sys{} pipelines (IntentInferrer disabled) produce 44 emergent behaviours: collaboration (32, 73\%), routine adaptation (7, 16\%), conflict resolution (4, 9\%), and group formation (1). SOTOPIA scores (Table~\ref{tab:sotopia}) show strength on \emph{goal} (9.5) and \emph{believability} (9.0) but weakness on \emph{social\_rules} ($-2.13$) and \emph{secret} ($-0.67$), indicating that structured environment access supports local coordination but not higher-order social constraint handling. These data supplement the SOTOPIA evaluation; they do not constitute a separate theory-of-mind claim.

\end{document}